\documentclass[10pt,twocolumn,letterpaper]{article}

\usepackage{cvpr}              

\usepackage{times}
\usepackage{graphicx}
\usepackage{amsmath}
\usepackage{amssymb}
\usepackage{pifont}
\usepackage{array}
\usepackage{comment} 
\usepackage{capt-of}
\usepackage{booktabs}
\usepackage{tabularx}
\usepackage{soul}
\usepackage{multirow}
\usepackage{multicol}
\usepackage[dvipsnames]{xcolor}
\usepackage{xhfill}
\usepackage{float}
\usepackage{textcomp}
\usepackage{enumitem}

\definecolor{cvprblue}{rgb}{0.21,0.49,0.74}
\usepackage[pagebackref,breaklinks,colorlinks,allcolors=cvprblue]{hyperref}

\def\paperID{3} 
\def\confName{CVPR}
\def\confYear{2025}

\title{What Makes for a Good Stereoscopic Image?}

\author{\vspace{1mm} Netanel Y. Tamir$^{*\,1,2}$ \hspace{2mm} Shir Amir$^{*1}$ \hspace{2mm} Ranel Itzhaky$^{1}$ \hspace{2mm} Noam Atia$^{1}$ \\ \vspace{1mm} Shobhita Sundaram$^{3}$ \hspace{2mm} Stephanie Fu \hspace{2mm} Ron Sokolovsky$^{1}$ \hspace{2mm} Phillip Isola$^{3}$ \\ \vspace{3mm} Tali Dekel$^{2}$ \hspace{2mm} Richard Zhang \hspace{2mm} Miriam Farber$^{1}$ \\
\small$^{1}$Apple \hspace{2mm} $^{2}$Weizmann Institute of Science \hspace{2mm} $^{3}$MIT \\
}
\begin{document}
\twocolumn[{ %
\vspace{-2.9em}
\maketitle

\renewcommand\twocolumn[1][]{#1}%
\vspace{-2.5em}

\begin{center}
    \centering
    \setlength{\abovecaptionskip}{0cm}
    \setlength{\belowcaptionskip}{0.5cm}
    \includegraphics[width=0.95\linewidth]{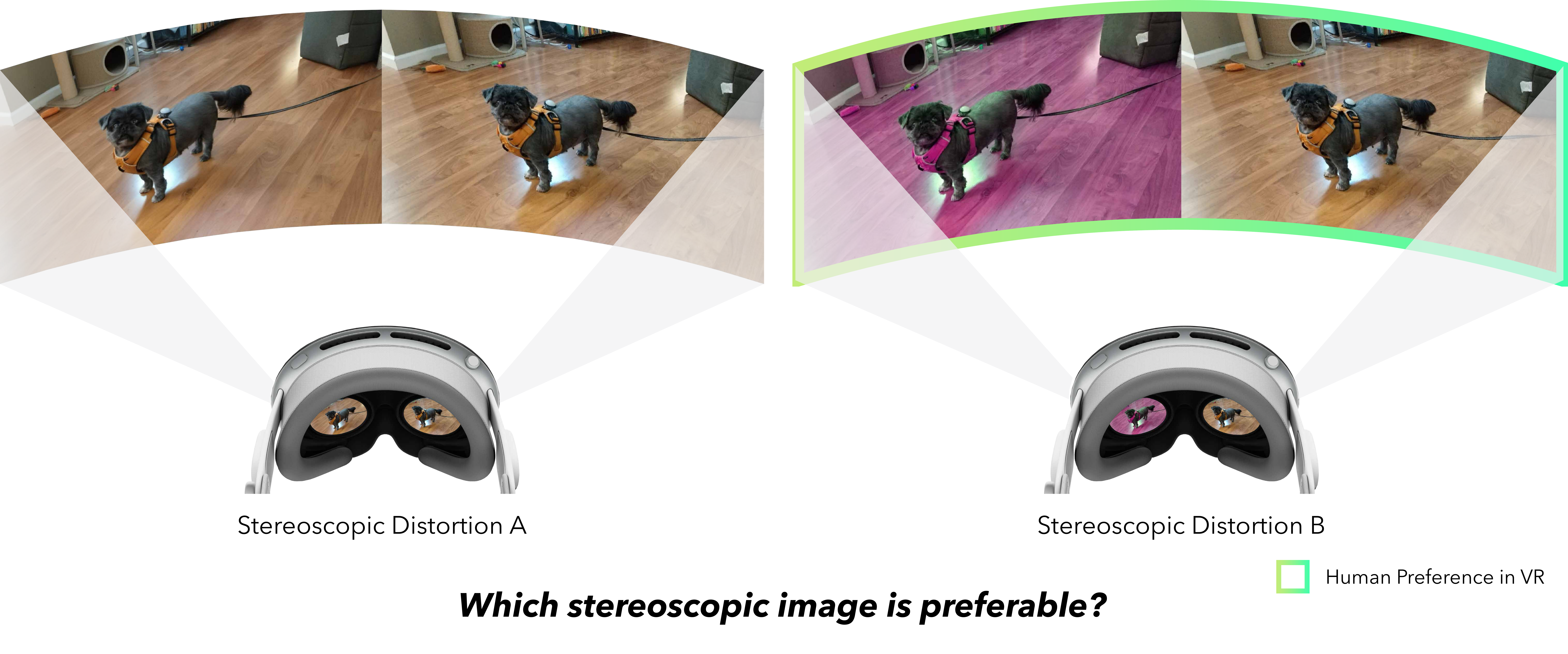}
    \captionsetup{type=figure}\caption{\small{\bf Capturing human preference in VR.} 
    We present a stereoscopic dataset created by showing participants two distorted versions of a stereoscopic image through a VR headset and asking which one they prefer. Our dataset encompasses a variety of distortion types. In the example above, the left image of each stereo pair is distorted: (left stereo image) SDEdit~\cite{meng2021sdedit} is applied resulting in texture changes, such as on the floor and dog; (right stereo image) a shift in hue is used as the distortion. Interestingly, we find that preferences in VR often differ from preferences on screen. We leverage this dataset to train a stereo quality prediction model.
    }
    \label{fig:teaser}
\end{center}%
}]

\begin{abstract}
With rapid advancements in virtual reality (VR) headsets, effectively measuring stereoscopic quality of experience (SQoE) has become essential for delivering immersive and comfortable 3D experiences. However, most existing stereo metrics focus on isolated aspects of the viewing experience such as visual discomfort or image quality, and have traditionally faced data limitations. To address these gaps, we present SCOPE (Stereoscopic COntent Preference Evaluation), a new dataset comprised of real and synthetic stereoscopic images featuring a wide range of common perceptual distortions and artifacts. The dataset is labeled with preference annotations collected on a VR headset, with our findings indicating a notable degree of consistency in user preferences across different headsets. Additionally, we present iSQoE, a new model for stereo quality of experience assessment trained on our dataset. We show that iSQoE aligns better with human preferences than existing methods when comparing mono-to-stereo conversion methods.
\end{abstract}    
\vspace{-5mm}
\section{Introduction}
\label{sec:intro}

\textit{``What is to come of the stereoscope and the photograph we are almost afraid to guess, lest we should seem extravagant.'' -- Oliver Wendell Holmes, 1859}

Stereoscopy, commonly referred to as stereo imaging, is a technique in which a slightly varied image is displayed separately to each eye, creating the illusion of a 3D scene. For over a century, this method has been used with a diverse array of viewing tools, starting from the revolutionary Victorian-era stereoscopes, to the red and cyan anaglyph glasses popularized in the $20\textsuperscript{th}$ century~\cite{dubois2001anaglyph}, and now rapidly advancing virtual reality (VR) headsets. 
With recent improvements in the quality of VR headsets, stereoscopic image quality is more important than ever to evaluate. Various factors may effect the final quality, including 2D aesthetics, depth sensation (stereopsis) and viewing comfort; making this task particularly challenging. While stereo imaging has a long history, the current tools available for assessing these images are still quite limited. 

Recent advancements in neural rendering have paved the way for advancing immersive 3D content creation.
Neural Radiance Fields (NeRFs) \cite{mildenhall2020nerf,barron2021mipnerf,barron2022mip,barron2023zipnerf,duckworth2023smerf} and 3D Gaussian Splats (3DGS) \cite{kerbl20233d, fan2024instantsplat, yu2024mip}, in particular, have emerged as powerful scene representation techniques, allowing for efficient and realistic scene rendering. 
In addition, many text-to-3D models leverage recent progress in text-to-image generation 
\cite{jain2022zero,poole2022dreamfusion,lin2023magic3d,wang2024prolificdreamer,yi2024gaussiandreamer}, often using 2D diffusion priors to generate realistic 3D content. 
Moreover, the recent boom in text-to-video generation \cite{wang2023modelscope,bar2024lumiere, videoworldsimulators2024,yang2024cogvideox} has paved the way for advances in stereoscopic video synthesis. All of these inspired recent stereoscopic images and video generation algorithms~\cite{dai2024svg,zhao2024stereocrafter,lv2024spatialdreamer,shi2024immersepro}.

This progress in 3D content creation, which serves as a source for stereo content, is occurring alongside improvements in stereo image capturing technology.
Dual or multi-camera systems are standard in today's smartphones, providing the necessary hardware for stereo photography \cite{counterpointresearchDualCamera,treendlyCapturingFuture}.
Some models leverage these multiple cameras to enable capturing true stereo images \cite{appleStereoVideo,dxomarkHYDROGENHolographic}.
In parallel, VR headsets have become more accessible than ever, with millions of users and a multi-billion dollar market \cite{NRG2022beyond, fortunebusinessinsightsVirtualReality}.
These complementary trends lead us to a moment ripe for reassessing our methods for evaluating stereo imagery.

Despite the surge in stereo content, evaluation methods for it remain limited. While single-image quality assessment (IQA) tools exist \cite{mittal2012no, yang2022maniqa, wang2022exploring, wu2023q}, they do not capture the complex relations between the two monocular images of a stereo pair, such as viewing comfort or realistic depth, when viewed together in stereoscopic 3D. On the other hand, existing methods for evaluating stereo content suffer from a lack of annotated training data \cite{oh2018deep,si2022no,zhang2023towards,zhou2023stereoscopic}, and focus predominantly on a few low-level artifacts (e.g. noising and excessive horizontal disparity). Furthermore, these methods have not been trained or tested on more complex artifacts (e.g. results of inpainting or depth algorithms).

To address these bottlenecks, we present a new benchmark for stereo imagery. Our dataset, SCOPE -- Stereoscopic COntent Preference Evaluation -- is comprised of pairs of distorted stereo images. Some of these images are created by distorting physically-captured stereo images, while others are created via generative methods applied to monocular images. The distortions are selected from a wide range of common image augmentations (e.g. photometric and spatial distortions) and generative methods including 3D Gaussian splatting \cite{kerbl20233d} and MotionCtrl \cite{wang2023motionctrl}, to encompass the variety of artifacts that may appear when generating images. Human annotators then participate in a two-alternative forced choice (2AFC) test, where they vote on which version of a given stereo image they prefer when viewed through a VR headset. Using this dataset, we train a stereo quality of experience (SQoE) assessment model, meant to capture a holistic sense of the overall visual experience. We demonstrate its practicality in assessing different mono-to-stereo generation methods and its ability to extrapolate to both distortion types and strengths that are not present in its training data. To the best of our knowledge, this is the largest data-driven effort to develop an SQoE evaluator, addressing a broad range of stereo artifacts.

To summarize, our key contributions are as follows:
\begin{itemize}[topsep=0pt,itemsep=0pt]
\item SCOPE - A two-alternative forced choice (2AFC) stereoscopic dataset, containing $2400$ samples annotated by $103$ participants;
\item iSQoE - An SQoE model trained using our dataset;
\item Demonstrating our model's effectiveness in assessing off-the-shelf stereo synthesis methods. 
\end{itemize}

\section{Related Work}
\label{sec:related_work}

\subsection{Stereo Image Assessment}
Stereo Quality of Experience (SQoE) encompasses the user's overall viewing experience of the stereoscopic 3D content. Compared to traditional monoscopic image assessment, stereoscopic evaluation presents unique challenges, as factors like visual discomfort and depth perception may influence a user's overall impression or satisfaction \cite{qi2012Quality, liu2019Learning}.

Some works evaluate SQoE through \textit{stereo image quality assessment} (SIQA), an extension of single image quality assessment, which aims to evaluate the degree of distortion in images.
No-reference quality assessment methods are widely seen as the most practical and adaptable, with many no-reference stereo image quality assessment (NR-SIQA) methods having been proposed over the years. These include both hand-crafted feature-based methods \cite{akhter2010no,chen2013no,shen2018no,li2019no,liu2020no} and deep learning-based techniques \cite{zhang2016learning, jia2018saliency,zhou2019dual,fang2019stereoscopic,shen2021no, si2022no,zhang2023towards}. 

Other studies have concentrated on evaluating \textit{discomfort} experienced when viewing stereoscopic images, which are not addressed by traditional image quality assessment methods. Several factors contributing to viewer discomfort have been highlighted in the literature \cite{ukai2008visual,shibata2011zone,yano2004two,kooi2004visual,bando2012visual,lambooij2009visual}. Visual discomfort predictors developed for this purpose include approaches based on hand-crafted features \cite{kim2011visual,park20143d,poulakos2015computational,chen2017visual} as well as deep learning-based models \cite{Kim2019BinocularFN,oh2018deep,zhou2023stereoscopic}. Like the aforementioned SIQA models, these models rely on small-scale datasets, limiting their overall effectiveness. 

Our work aims to capture the overall impression of a stereo image, not only image quality or viewer experience. We provide a single data-driven model that implicitly considers both image quality and user comfort based on user annotations.  

Additionally, there are IQA models developed specifically for VR (VR-IQA). These mainly focus on evaluating monoscopic and stereoscopic \textit{omnidirectional} images, commonly referred to as $360$\textdegree ~images \cite{lim2018vr, chen2019study, sui2021perceptual, poreddy2023no}. Evaluating such images presents unique challenges, such as projection distortions, that differ from the focus of our work.

\subsection{Stereoscopic Datasets}
Numerous stereoscopic image datasets are publicly accessible, offering a variety of sizes, resolutions, and camera baselines \cite{Butler:ECCV:2012,scharstein2014high,menze2015object,Flickr1024,hua2020holopix50k,Mehl2023_Spring}. The viewer's depth perception and overall comfort while viewing stereo images are affected by how closely the camera baseline matches their interpupillary distance, which averages approximately $63$ mm \cite{dodgson2004variation}. Several of these datasets include human-annotated preferences regarding quality or comfort, and as such were used to develop automated assessment methods \cite{moorthy2013subjective, park20143d, wang2014quality, wang2015quality}. However, the limited size and diversity of these annotated datasets has become the main bottleneck in advancing automated assessment techniques.

\subsection{Psychophysics in Virtual Reality}
Numerous psychophysical studies have investigated human perception and cognition within VR environments to understand how users experience and interact with these systems. Krajancich \etal \cite{10.1145/3414685.3417820} demonstrated that adjusting rendering based on user gaze enhances depth perception and realism, while Guan \etal \cite{guan2023perceptual} highlighted users' sensitivity to small errors in camera positioning. Additionally, work by Chen \etal \cite{chen2024pea} examined the effects of power-saving techniques on perceptual quality in VR, and Matsuda \etal \cite{matsuda2022realistic} found that current VR headsets often fall short of user expectations for display brightness. Thomas \cite{thomas2020examining} found that users can accurately perceive the size of virtual objects within arm’s reach in VR, with height and width judgments closely matching actual dimensions. Collectively, this body of research underscores the critical role of understanding human perception to guide the development of effective and user-centric VR technologies. In our work we implicitly capture characteristics of human stereoscopic perception within our dataset, and train a model to predict SQoE accordingly.

\begin{table}[t]
    \begin{center}
    \begin{tabular}{ r p{4cm} }
    \toprule
    \textbf{Sub-type} & \textbf{Distortion type} \\ 
    \midrule
    \multirow{2}{*}{Novel-view synthesis} & 2D Lifting, MotionCtrl, \\ & 3D Gaussian splatting \\
    \midrule
    \multirow{3}{*}{Noise} & Uniform white noise, Gaussian white noise, Checkerboard artifact \\
    \midrule
    Blur & Average blur, Gaussian blur \\
    \midrule
    Compression & JPEG compression \\
    \midrule
    \multirow{3}{*}{Photometric} & Hue shift, Saturation shift, \\ & Brightness shift, Contrast shift\\
    \midrule
    \multirow{3}{*}{Spatial} & Magnification, Rotation, Keystone effect, Warping, Chromatic aberration\\
    \midrule
    Diffusion-based editing & SDEdit \\
    \bottomrule
    \end{tabular}
    \caption{\small \textbf{Distortion Pool.} The diverse set of distortions applied to stereoscopic images generates a wide variety of artifacts. 
    }
    \label{tab:augmentations}   
    \end{center}
    \vspace{-6mm}
\end{table}
\section{Stereoscopic Dataset Collection \& Learning}
\label{sec:stereoscopic_dataset_collection}

\subsection{SCOPE Dataset}
\label{sec:data_creation}

We collect human preferences for stereoscopic 3D experiences by creating different variations of stereo images. Each stereo image undergoes two distinct distortions among those listed in \Cref{tab:augmentations}. The stereo images are then compared by five annotators specifying which version they prefer. 

\begin{figure*}[ht]
    \centering
    \includegraphics[width=0.82\linewidth]{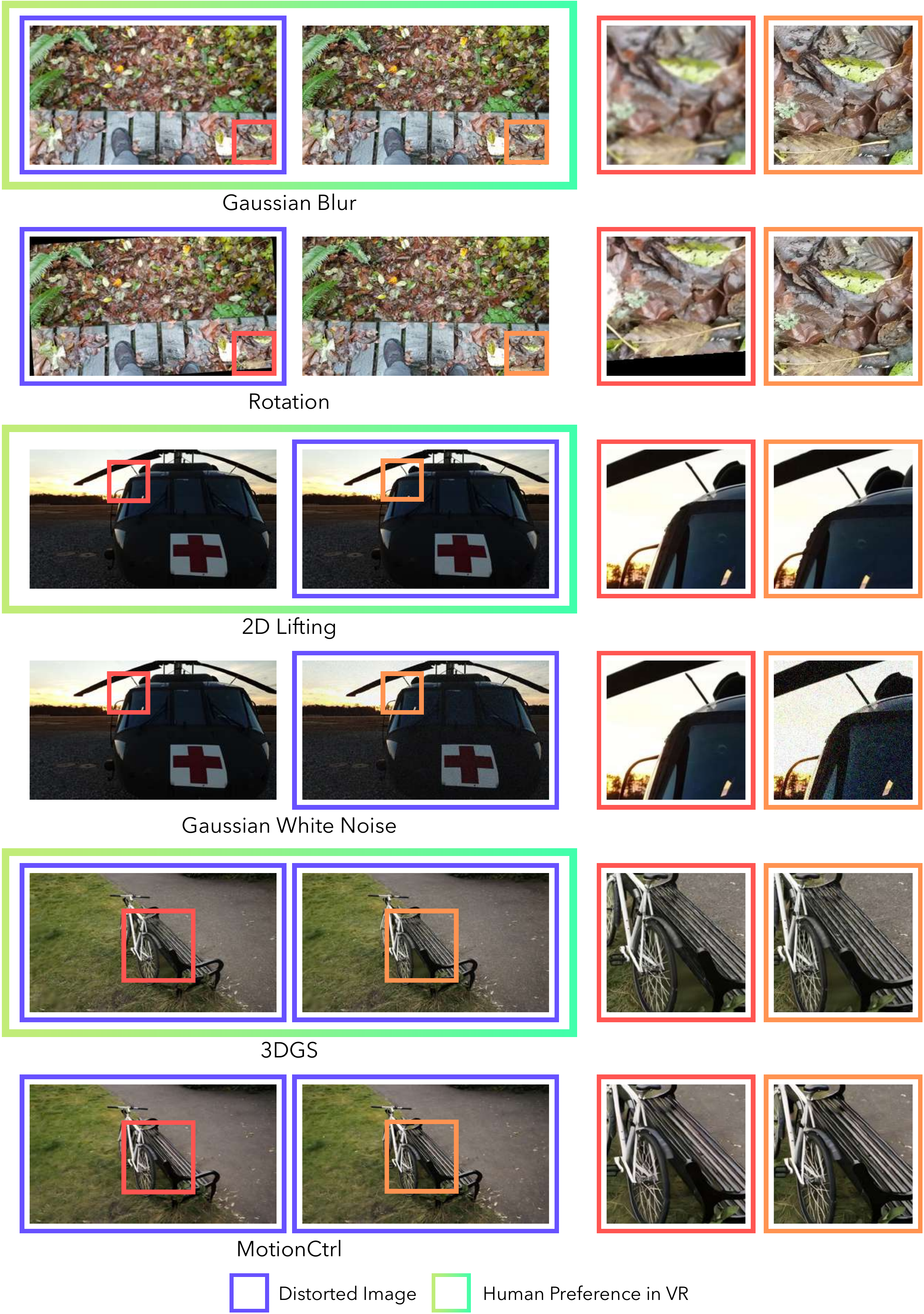}
    \caption{\small
    {\bf Dataset examples.} Each stereo image was subjected to two different distortions, applied consistently to either the left, right, or both images. Participants in a VR-based user study were then asked to choose their preferred version. 
    On the right of each sample, we zoom to highlight the differences between the images. Some distortions are more easily visible in 2D (\eg Gaussian White Noise, Rotation) while others are more visible on VR devices (\eg disparity differences cause increased depth sensation in the 2D lifting example).
    }
    \label{fig:3_examples}
\end{figure*}

\begin{figure*}[t]
    \centering
    \includegraphics[width=0.90\linewidth]{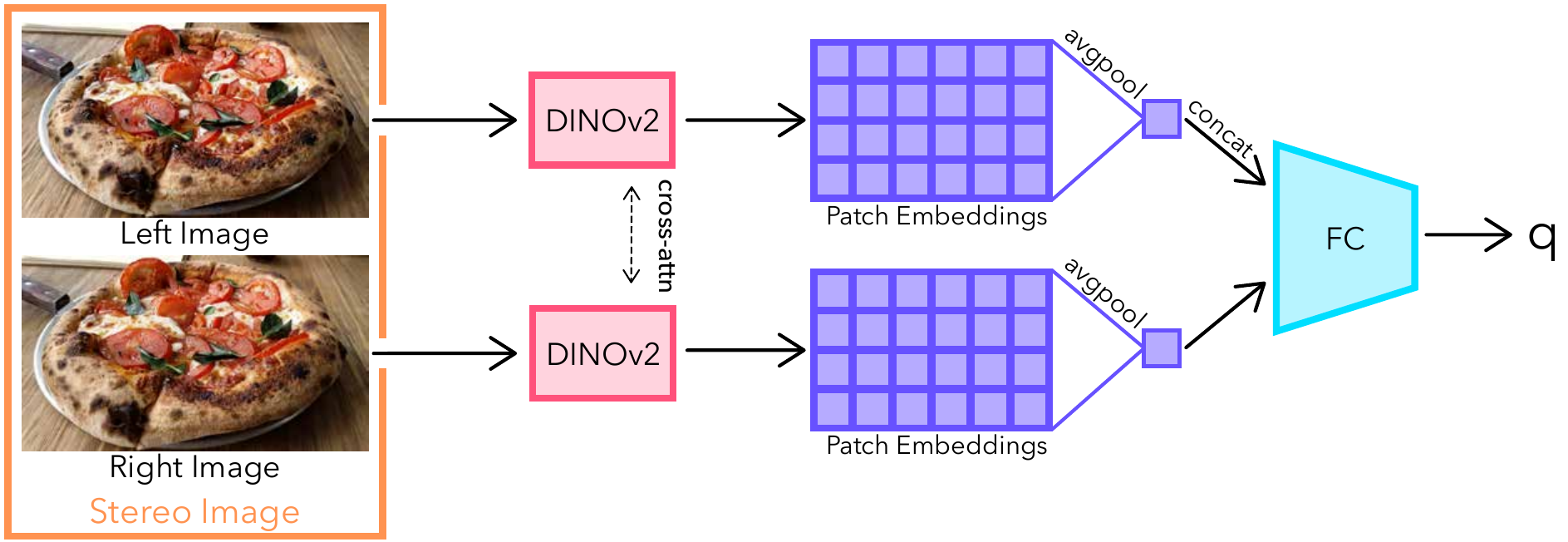}
    \caption{\small
    {\bf Model architecture.} The left and right images of a stereo pair are processed by a modified DINOv2~\cite{oquab2023dinov2} network with cross-attention between images. The resulting spatial tokens are pooled, concatenated, and passed through a small fully-connected network, outputting a single value indicating quality (lower is better). We train the model with a hinge loss and LoRA \cite{hu2021lora} for the DINOv2 network.
    }
    \vspace{-0.3cm}
    \label{fig:model}
\end{figure*}

\noindent Each stereo image in SCOPE is created in one of the following manners: 
\begin{enumerate}[label=(\roman*), leftmargin=15pt]
    \item \textbf{Single image to stereoscopic image.} We use one of two distinct approaches: (a) MotionCtrl \cite{wang2023motionctrl}, an image-to-video model which enables camera control, while keeping the scene generally static. We simulate interocular distance by imposing horizontal camera motion to the input images. (b) we create a \textit{2D lifting} pipeline, using one of several off-the-shelf monocular depth estimators \cite{Ranftl2022,ke2023repurposing,yang2024depth} to estimate disparity and forward warp the image accordingly. Then, we use LaMa inpainting~\cite{suvorov2022resolution} to fill in the dis-occluded regions. 
    \item \textbf{Multiview images to stereoscopic image.} We use 3D Gaussian splatting~\cite{kerbl20233d} fitted to multi-view scenes from several datasets~\cite{Knapitsch2017,DeepBlending2018,barron2022mip}. We rendered each stereo image as two 2D images with a horizontal offset between them. For each scene we used two different 3DGS representations optimized at different levels, resulting in differing amounts of artifacts.
    \begin{sloppypar}
    \item \textbf{Distorting \textit{existing} stereoscopic images.} We leverage physically acquired stereoscopic images from Holopix50k HD~\cite{hua2020holopix50k} and apply a diverse set of image editing techniques as distortions. These include noise injection, photometric and spatial transformations, and SDEdit~\cite{meng2021sdedit} to modify the high-frequency components of the images.
    \end{sloppypar}
\end{enumerate}

\noindent The dataset is hence comprised of two subsets to avoid accumulating distortions: The first uses distortion types (i) and (iii) on images from Holopix50k HD~\cite{hua2020holopix50k}, which contains various in-the-wild physically captured stereo images. The second applies distortion types (i) and (ii), and is based on $13$ multi-view scenes from several datasets: Tanks and Temples~\cite{Knapitsch2017}, Deep Blending~\cite{DeepBlending2018} and Mip-Nerf~\cite{barron2022mip}.

We create $2400$ data samples of resolution $1280 \times 720$, $2000$ examples from the first subset and $400$ from the second. \Cref{fig:teaser,fig:3_examples} contain dataset examples. Each unique image, drawn from existing datasets, serves as the basis for two distorted versions, and is used only once. For every data sample we ensure the distortions are applied consistently on either the left, right, or both images uniformly, reducing noise caused by ocular dominance \cite{porac1976dominant} (see \Cref{fig:3_examples}). The variety of distortion types results in a diverse array of artifacts which can lead to viewer discomfort, reduce image quality, and/or alter depth perception. More information is available in \Cref{sec:dataset_breakdown} of the supplementary material (SM). \Cref{tab:stereoscopic_datasets} of the SM compares our dataset to previous stereoscopic preference datasets, indicating it is has significantly more samples than previous datasets. Furthermore, it is the first dataset with 2AFC annotations, that can be directly used to train a SQoE model.

\subsection{VR Annotations in SCOPE}

We conduct a large scale 2AFC study, similar to LPIPS \cite{zhang2018unreasonable} and DreamSim \cite{fu2023dreamsim} where participants are asked to annotate our dataset from \Cref{sec:data_creation} on an Apple Vision Pro headset.
Specifically, for each real stereoscopic image pair $x = (x_l,x_r)$, users compare two modified versions. These versions are defined as ${(x^m_{l},x^m_{r}),(x^n_{l},x^n_{r})}$, where $l$ and $r$ are left and right views in a stereo image respectively; and $m$ and $n$ are distinct distortions from \Cref{tab:augmentations}. These are applied to at least one view in each stereo pair, and are applied to the same views across both stereo pairs.

We collect judgments $y\in \{m,n\}$ by asking participants which version is \textit{preferable}, as depicted in \Cref{fig:teaser}. We pose the question to the participants in this manner to collect majority-based binary labels, which can then be used to train a holistic SQoE model. We also ask participants to avoid closing one eye at a time and instead view the left and right images simultaneously. The user study was conducted in batches of $25$ examples, with breaks between batches to mitigate visual fatigue, a condition characterized by Lambooij \etal.~\cite{lambooij2009visual} as eye strain and reduced visual performance after prolonged viewing. 
Our dataset, ${D = \bigl\{\bigl((x^m_{l},x^m_{r}),(x^n_{l},x^n_{r})\bigr),y\bigr\}}$, ultimately contains $2400$ examples, each labeled with $5$ annotations and collected from $103$ participants. We find that a third of the examples are unanimously agreed upon, with a $5/0$ split, another third have a $4/1$ split, and the remaining third have a $3/2$ split. We randomly partition our data into train ($80\%$), validation ($10\%$), and test ($10\%$) sets. We name our dataset SCOPE -- Stereoscopic COntent Preference Evaluation -- and make it publicly available.

\begin{table*}[ht]
\begin{center}
\resizebox{1.\linewidth}{!}{
\begin{tabular}{ccccc cccc}
 \toprule
 \multicolumn{5}{c}{\bf Model} & \multicolumn{4}{c}{\bf Mean Accuracy}\\
 \cmidrule(lr){1-5}\cmidrule(lr){6-9}
 {\bf Ablation type} & {\bf Resolution} & {\bf Backbone} & {\bf Attention Fusion [Layers]} & {\bf LoRA} & {\bf 3--2 Split} & {\bf 4--1 Split} & {\bf 5--0 Split} & {\bf Total}\\
 \midrule
 \multirow{1}{*}{\shortstack[c]{\bf Image Resolution}} & 
 $224 \times 224$ & DINOv2 S/14~\cite{oquab2023dinov2} & Concat $[2,5,8,11]$ & \checkmark & 60.4 & 68.4 & 83.5 & 70.8 \\
 \midrule
 \multirow{4}{*}{\shortstack[c]{\bf Backbone}} 
 & $1280 \times 720$  &  CLIP L/14~\cite{radford2021learning} & Concat $[2,5,8,11]$ & \checkmark & 59.5 & 66.7 & 78.2 & 68.2 \\
 & $1280 \times 720$  & OpenCLIP L/14~\cite{ilharco_gabriel_2021_5143773} & Concat $[2,5,8,11]$ & \checkmark & 61.4 & 68.1 & 78.4 & 69.3 \\
 & $1280 \times 720$  & Croco~\cite{weinzaepfel2023crocov2improvedcrossview} & Unmodified & Dec. & 59.0 & 67.5 & 80.1 & 69.0 \\
 & $1280 \times 720$  & StereoQA-Net~\cite{zhou2019dual} & -- & -- & 62.5 & 64.6 & 77.3 & 68.1 \\

 \midrule
 \multirow{4}{*}{\shortstack[c]{\bf Attention \\\textbf{Module}}} & 
 $1280 \times 720$  &  DINOv2 S/14 & Swap $[2,5,8,11]$ & \checkmark & \textbf{62.6} & \textbf{72.2} & 83.6 & 73.0 \\
 & $1280 \times 720$  & DINOv2 S/14 & Concat $[0-11]$ & \checkmark & 61.7 & 69.3 & 84.4 & 71.8 \\
 & $1280 \times 720$  & DINOv2 S/14 & Concat $[11]$ & \checkmark & 60.2 & 72.1 & 83.9 & 72.2 \\
 & $1280 \times 720$  & DINOv2 S/14 & Unmodified & \checkmark & 60.6 & 71.7 & 82.6 & 71.9 \\
 \midrule

 \multirow{1}{*}{\shortstack[c]{\bf Optimization}} & 
 $1280 \times 720$ & DINOv2 S/14 & Concat $[2,5,8,11]$ & -- & 60.4 & 68.7 & 82.2 & 70.5 \\
 \midrule
 
 \multirow{1}{*}{\shortstack[c]{\bf iSQoE (Ours)}} & 
 $1280 \times 720$ & DINOv2 S/14 & Concat $[2,5,8,11]$ & \checkmark & 62.1 & 72.0 & \textbf{84.8} & \textbf{73.1} \\

 \bottomrule
\vspace{-5mm}
\end{tabular}
}
\end{center}
\caption{\small {\bf Model ablations.} Our chosen variant performs the best on the entire test set as well as the 5-0 uniform split, with the other attention fusion variants being close.}\label{tab:ablations}
\vspace{-3mm}
\end{table*}

\subsection{Training an SQoE Model}

After collecting our dataset, we use it to train an SQoE model which given a stereoscopic image as input, can then produce a single score that takes into account quality, comfort, and depth sensation. We name our model iSQoE - immersive Stereoscopic Quality of Experience assessor. The architecture is described in \Cref{fig:model} and is inspired by LPIPS~\cite{zhang2018unreasonable} and DreamSim~\cite{fu2023dreamsim} which also train perceptual models on 2AFC annotations. 

Since stereoscopic images are composed of two 2D images, we leverage existing pretrained image backbones to process each image in a stereoscopic pair. The resulted features are pooled, concatenated, and passed through a lightweight fully-connected network, which generates a quality score normalized by a sigmoid function. We train the model in a Siamese fashion \cite{bromley1993signature} with a hinge loss between the two stereo pairs using the participants' preferences as ground truth. To enable information flow between the 2D backbones of each image in the stereo pair we pass information between the attention modules of the pretrained backbones.

We conduct several ablations presented in \Cref{tab:ablations}. We report total mean accuracy, as well as on subsets with unanimous (5-0), majority (4-1) and ambiguous (3-2) annotations.  We consider several backbone features, such as those extracted from foundation models like DINOv2~\cite{oquab2023dinov2} and CLIP~\cite{radford2021learning}, and also backbones with task specific knowledge like Croco~\cite{weinzaepfel2023crocov2improvedcrossview} which was trained for Novel View Synthesis and StereoQA-Net~\cite{zhou2019dual} trained for SIQA. We found DINOv2 to yield the best results, with Croco not falling much behind, indicating 2D image understanding is important for SQoE. Similar to DreamSim we found that finetuning DINOv2 with LoRA yields an additional improvement.

Furthermore, we examine early information fusion between the 2D image backbones by manipulating the attention layers in the feature extractors, to enable information sharing within the left and right parts of the stereo image. We considered two fusion strategies: (i) \textit{swapping} the keys and values between left and right images and (ii) \textit{concatenating} the keys and values from left and right images. Queries are kept in place in both methods. Inspired by Croco~\cite{weinzaepfel2023crocoselfsupervisedpretraining3d} we examined fusion on all layers [0-11], last layer [11], and alternating fusing and non-fusing layers [2, 5, 8, 11]. We found \textit{concatenating} fusion on alternating layers to perform best, and all fusion forms to perform better than no fusion at all. Intuitively, this could imply sharing information across left and right images is beneficial for stereo understanding. 

\begin{figure}[ht]
    \centering
    \includegraphics[width=0.95\linewidth]{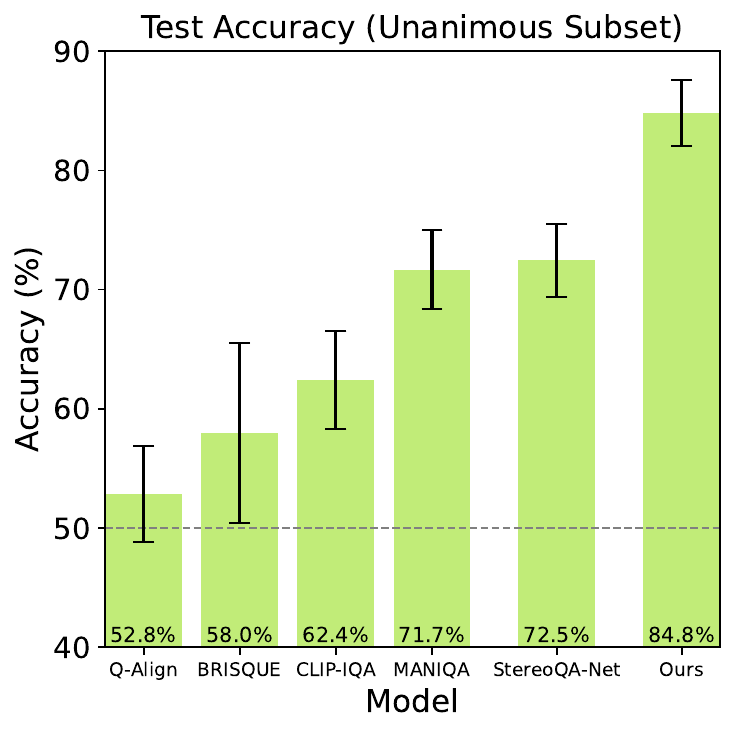}
    \caption{\small{\bf Test accuracy on SCOPE.} We report the mean and standard deviation of the unanimous cases in the test set over several splits. Our model outperforms the other SIQA and IQA models.}
    \vspace{-0.6cm}
    \label{fig:mini_performance}
\end{figure}

\begin{figure*}[ht]
    \centering
    \includegraphics[width=0.89\linewidth]{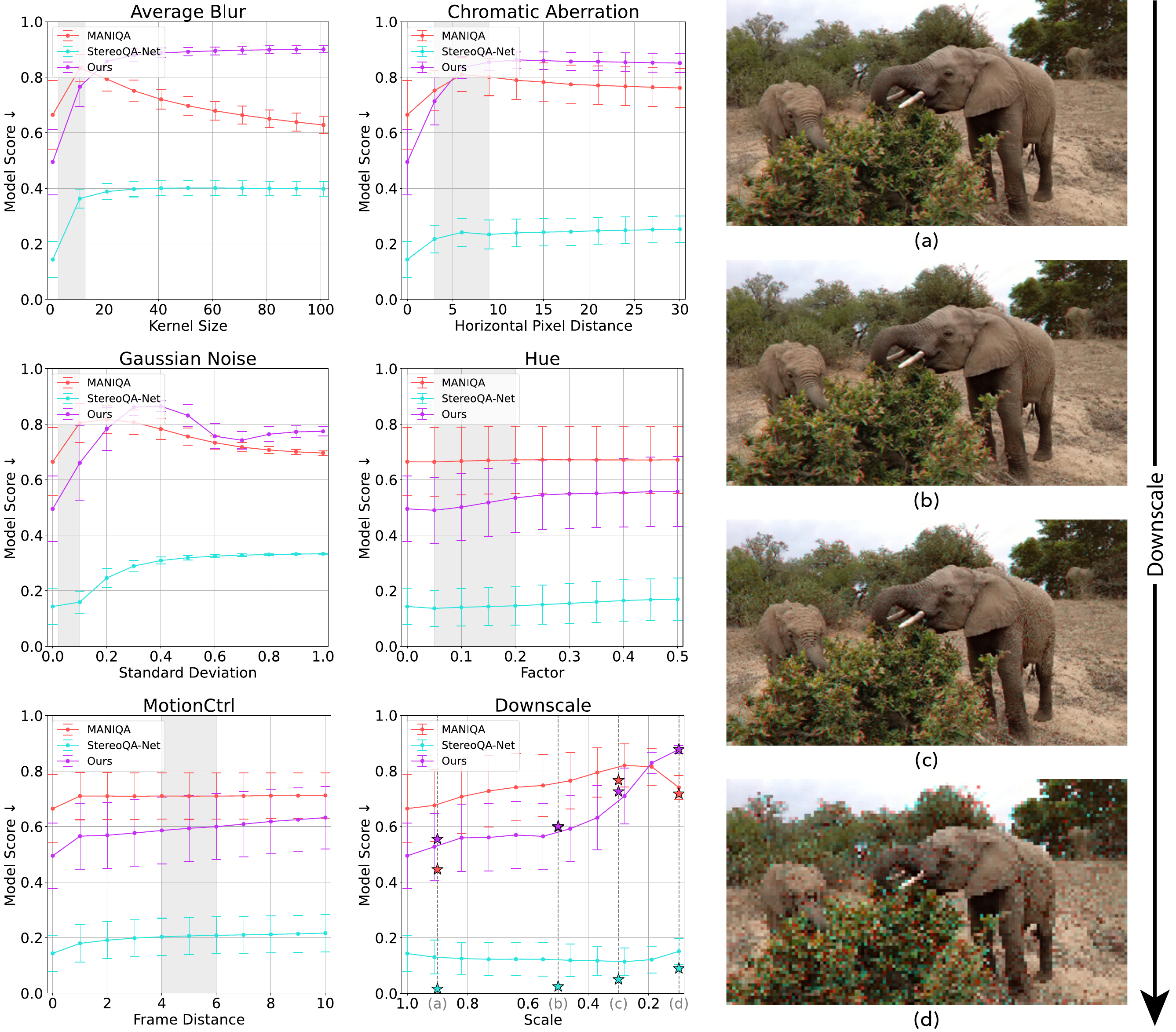}
    \caption{\small{\bf Progressive degradation evaluation.} We report model scores on $200$ stereo images for six different distortions. The gray regions indicate the distortion intensity used in SCOPE. Stereo images (a) -- (d), presented as anaglyphs, exhibit progressive downscaling and are represented by stars.}
    \vspace{-0.2cm}
    \label{fig:aug_strength}
\end{figure*}
\section{Results \& Analysis}
\label{sec:results}

\subsection{Performance on the SCOPE Dataset}

We benchmark several models on SCOPE in \Cref{fig:mini_performance}. We consider the known IQA metrics BRISQUE~\cite{mittal2012no}, MANIQA~\cite{yang2022maniqa}, CLIP-IQA~\cite{wang2022exploring} and Q-Align~\cite{wu2023q}. For each stereo image we calculate the IQA score for the left and right views individually and use their mean as the score for the entire stereo image. We also consider StereoQA-Net~\cite{zhou2019dual}, an existing NR-SIQA model, based on a CNN \cite{lecun1998gradient,krizhevsky2012imagenet} architecture and trained on the LIVE 3D Phase I dataset \cite{moorthy2013subjective}. Further comparisons are limited due to availability of code or pretrained models.

We partition the dataset into train, validation, and test subsets several times. When evaluating model performance, we report the accuracy specifically on the subset of the test data that has unanimous human annotations. These unanimously-annotated examples are likely to contain the least amount of annotation noise and represent the most cognitively impenetrable cases \cite{cavanagh1999cognitive,stokes2013cognitive}. \Cref{fig:mini_performance} shows our model successfully outperforms the other methods by a margin. Moreover, we find that MANIQA performs best within the IQA metrics, corroborating results from Zhu \etal.~\cite{zhu2024esiqa}. Additional information is available in \Cref{sec:sm_performance,sec:siqa_evaluation} of the SM.

\subsection{How do models respond to unseen distortions with varying strengths?}
\label{sec:aug_strength}
In \Cref{fig:aug_strength} we apply a series of degradations to stereo images with increasing severity and evaluate the distorted images using different models. The selected distortions may effect quality, comfort, and depth perception in VR. 
We also include downscaling, a distortion absent from SCOPE. We find that in all cases our model is able to extrapolate, behaving monotonically for a wider distortion range than it was trained on. We observe that MANIQA~\cite{yang2022maniqa} frequently rates an image as higher quality the more degraded it becomes, once an initial threshold is passed. StereoQA-Net~\cite{zhou2019dual} generally maintains monotonic behavior, however it tends to plateau quickly. Contrastingly, our model exhibits mostly monotonic behavior and a larger dynamic range, with gaussian noise being the primary exception. 

\subsection{How do viewing mediums influence stereo 3D perception?}
\label{sec:cross_device}

\begin{figure}[t]
    \centering
    \includegraphics[width=0.92\linewidth]{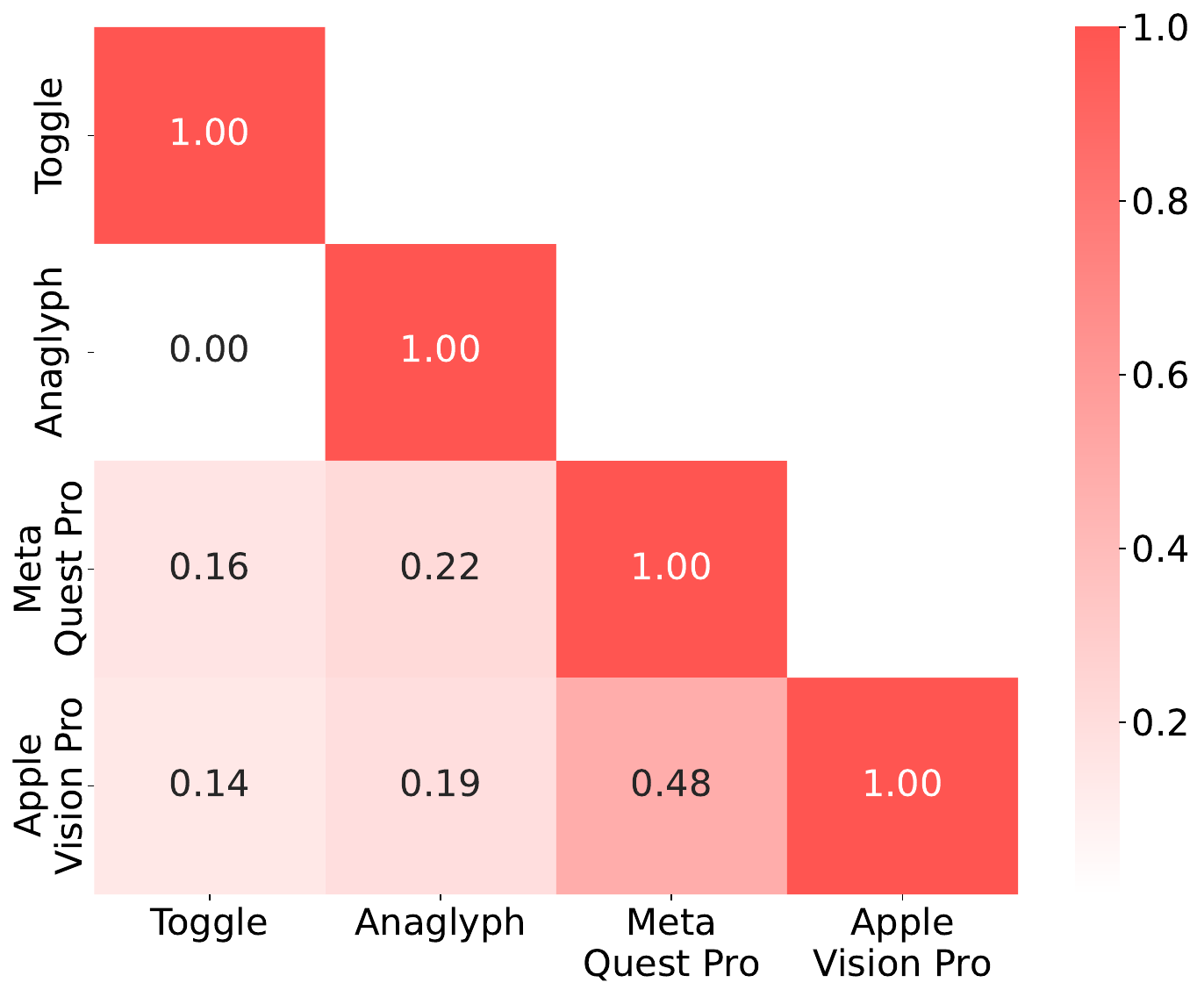}
    \caption{\small{\bf Viewing medium comparison.} We measure the correlation between human preferences across viewing mediums by calculating Cohen's kappa coefficient averaged across participants.}
    \vspace{-0.3cm}
    \label{fig:kappa}
\end{figure}

The underlying assumption that led us to collect annotations for SCOPE on VR, rather than a 2D screen, is that there is limited correlation between 2D perception of stereo images and their perception on VR devices. In \Cref{fig:kappa} we validate this hypothesis by randomly selecting $50$ samples from SCOPE and having $10$ participants repeatedly annotate them using different mediums: Apple Vision Pro, Meta Quest Pro, anaglyph images, and toggling between the left and right images on a monitor. We compute Cohen's kappa coefficient for each participant's responses across different mediums, then average these values across participants. 

Evidently, the correlation between VR devices (Meta Quest Pro, Apple Vision Pro) and non-VR devices (anaglyph, toggle) is quite low. Therefore, these simplified setups do not effectively reflect user experience on VR headsets, and thus are less suitable for SQoE data annotation. On the other hand, human preferences on Apple Vision Pro and Meta Quest Pro are have non-negligible correlation, which is a positive indication for the SCOPE's generalization across VR devices. In addition, we find that there is nontrivial agreement between participants across any specific viewing medium, as shown in \Cref{fig:agreement} in the SM. Notably, due to the cognitively penetrable nature of some examples in our dataset, perfect correlation would not be expected even for responses from the same user on the same device across two separate annotation sessions. See \Cref{sec:cross_device_sm} of the SM for more details. 

\subsection{Do models \& humans evaluate off-the-shelf stereo generators similarly?}

\begin{figure}[ht]
    \centering
    \includegraphics[width=0.92\linewidth]{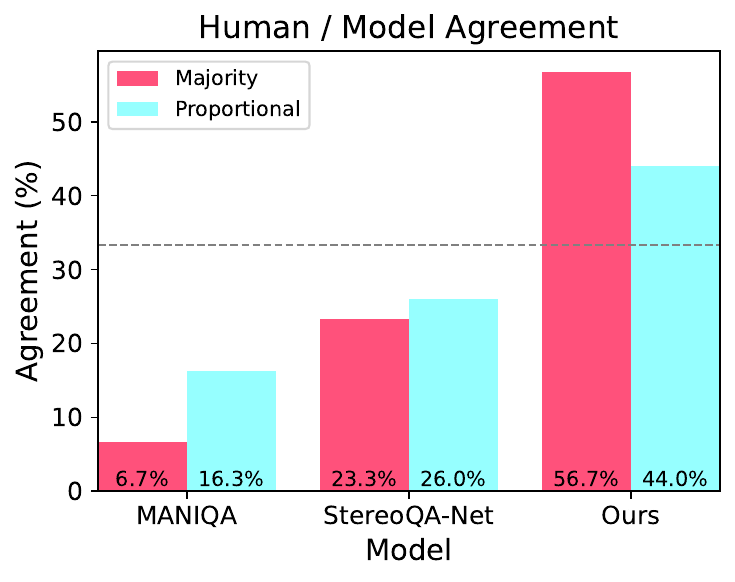}
      \caption{\small{\bf Alignment of stereo image preference between our participants and models.} We calculate the agreement in two different ways. (i) \textit{Majority}, where the model receives a binary score based on agreement with the majority human vote. (ii) \textit{Proportional}, in which the model score is proportional to the fraction of human votes for its preferred image.}
     \vspace{-0.3cm}
    \label{fig:alignment_exp}
\end{figure}

To assess our model's practicality for real use cases, we correlate human opinion with model ranking on off-the-shelf mono-to-stereo conversion techniques (see \Cref{fig:alignment_exp}): Depthify.ai \cite{depthifyDepthifyai}, Immersity AI \cite{immersity}, and Owl3D \cite{owl3dOwl3D}. We applied these to $30$ monoscopic images from the Spring dataset~\cite{Mehl2023_Spring}, with the main resulting artifacts being inaccurate depth, flawed inpainting, and jagged foreground edges. Importantly, our model did not see any images from Spring during training, and its animated artistic style is out of SCOPE's distribution. 

We conduct a user study using the Apple Vision Pro headset with $10$ participants, asking them to select their preferred stereo image from the three options generated for each original image. To minimize bias, we randomize both the order of the stereo image triplets and the arrangement of images within each triplet for every participant. Our findings indicate that participants generally favored the stereo images produced by Immersity AI, followed by Owl3D. Subsequently, we compare the alignment of user preferences with the assessments from MANIQA~\cite{yang2022maniqa} - the top-performing IQA method, Stereo-IQA~\cite{zhou2019dual}, and our proposed model. \Cref{fig:alignment_exp} exhibits our model has the highest correlation with human opinion. More information about this study appears in \Cref{sec:mono-to-stereo-evaluation} of the SM.
\section{Discussion}
\label{sec:discussion}

A major challenge in SQoE evaluation is the scarcity of annotated data. To address this, we curated a large dataset and developed a novel SQoE predictor to capture stereoscopic quality nuances. We validated its effectiveness by comparing novel-view synthesis methods, demonstrating superior alignment with human preference over existing methods.

However, our work has some inherent limitations, mainly resulting from our data being sourced from few existing sources. $41\%$ of our stereo images originate from the Holopix50k-HD~\cite{hua2020holopix50k} dataset and are distorted with a non-NVS distortion. This dataset was primarily captured using the RED Hydrogen One smartphone, which limits the horizontal disparity between the left and right images and affects depth perception. Moreover, our NVS subset is comprised of $400$ images taken from only $13$ scenes. Furthermore, our model inherits certain limitations from DINOv2~\cite{oquab2023dinov2}, which we chose as our backbone.

Despite these limitations, we believe that our work represents a significant step forward in SQoE evaluation. As stereoscopic content increases in prevalence, robust automated assessment is increasingly essential. We hope that our contributions lay the groundwork for more sophisticated and accurate evaluation methods.

\clearpage

\paragraph{Acknowledgements}
We thank all the participants in the user studies presented in this paper. We thank Bahjat Kawar, Yaron Ostrovsky Berman, Dror Simon and Matan Hugi for their early feedback. This work was partially supported by the Sagol Weizmann-MIT Bridge Program and an NSF GFRP Fellowship to SS.
{
    \small
    \bibliographystyle{ieeenat_fullname}
    \bibliography{main}

\begin{thebibliography}{106}
\providecommand{\natexlab}[1]{#1}
\providecommand{\url}[1]{\texttt{#1}}
\expandafter\ifx\csname urlstyle\endcsname\relax
  \providecommand{\doi}[1]{doi: #1}\else
  \providecommand{\doi}{doi: \begingroup \urlstyle{rm}\Url}\fi

\bibitem[Akhter et~al.(2010)Akhter, Sazzad, Horita, and Baltes]{akhter2010no}
Roushain Akhter, ZM~Parvez Sazzad, Yuukou Horita, and Jacky Baltes.
\newblock No-reference stereoscopic image quality assessment.
\newblock In \emph{Stereoscopic Displays and Applications XXI}, pages 271--282. SPIE, 2010.

\bibitem[Apple(2023)]{appleStereoVideo}
Apple.
\newblock Apple introduces spatial video capture on iphone 15 pro.
\newblock \url{https://www.apple.com/il/newsroom/2023/12/apple-introduces-spatial-video-capture-on-iphone-15-pro/}, 2023.
\newblock [Accessed 16-10-2024].

\bibitem[Bando et~al.(2012)Bando, Iijima, and Yano]{bando2012visual}
Takehiko Bando, Atsuhiko Iijima, and Sumio Yano.
\newblock Visual fatigue caused by stereoscopic images and the search for the requirement to prevent them: A review.
\newblock \emph{Displays}, 33\penalty0 (2):\penalty0 76--83, 2012.

\bibitem[Bar-Tal et~al.(2024)Bar-Tal, Chefer, Tov, Herrmann, Paiss, Zada, Ephrat, Hur, Li, Michaeli, et~al.]{bar2024lumiere}
Omer Bar-Tal, Hila Chefer, Omer Tov, Charles Herrmann, Roni Paiss, Shiran Zada, Ariel Ephrat, Junhwa Hur, Yuanzhen Li, Tomer Michaeli, et~al.
\newblock Lumiere: A space-time diffusion model for video generation.
\newblock \emph{arXiv preprint arXiv:2401.12945}, 2024.

\bibitem[Barron et~al.(2021)Barron, Mildenhall, Tancik, Hedman, Martin-Brualla, and Srinivasan]{barron2021mipnerf}
Jonathan~T. Barron, Ben Mildenhall, Matthew Tancik, Peter Hedman, Ricardo Martin-Brualla, and Pratul~P. Srinivasan.
\newblock Mip-nerf: A multiscale representation for anti-aliasing neural radiance fields.
\newblock \emph{ICCV}, 2021.

\bibitem[Barron et~al.(2022)Barron, Mildenhall, Verbin, Srinivasan, and Hedman]{barron2022mip}
Jonathan~T Barron, Ben Mildenhall, Dor Verbin, Pratul~P Srinivasan, and Peter Hedman.
\newblock Mip-nerf 360: Unbounded anti-aliased neural radiance fields.
\newblock In \emph{CVPR}, 2022.

\bibitem[Barron et~al.(2023)Barron, Mildenhall, Verbin, Srinivasan, and Hedman]{barron2023zipnerf}
Jonathan~T. Barron, Ben Mildenhall, Dor Verbin, Pratul~P. Srinivasan, and Peter Hedman.
\newblock Zip-nerf: Anti-aliased grid-based neural radiance fields.
\newblock \emph{ICCV}, 2023.

\bibitem[Bromley et~al.(1993)Bromley, Guyon, LeCun, S{\"a}ckinger, and Shah]{bromley1993signature}
Jane Bromley, Isabelle Guyon, Yann LeCun, Eduard S{\"a}ckinger, and Roopak Shah.
\newblock Signature verification using a" siamese" time delay neural network.
\newblock \emph{NeurIPS}, 6, 1993.

\bibitem[Brooks et~al.(2024)Brooks, Peebles, Holmes, DePue, Guo, Jing, Schnurr, Taylor, Luhman, Luhman, Ng, Wang, and Ramesh]{videoworldsimulators2024}
Tim Brooks, Bill Peebles, Connor Holmes, Will DePue, Yufei Guo, Li Jing, David Schnurr, Joe Taylor, Troy Luhman, Eric Luhman, Clarence Ng, Ricky Wang, and Aditya Ramesh.
\newblock Video generation models as world simulators.
\newblock 2024.

\bibitem[Butler et~al.(2012)Butler, Wulff, Stanley, and Black]{Butler:ECCV:2012}
D.~J. Butler, J. Wulff, G.~B. Stanley, and M.~J. Black.
\newblock A naturalistic open source movie for optical flow evaluation.
\newblock In \emph{ECCV}, 2012.

\bibitem[Cavanagh(1999)]{cavanagh1999cognitive}
Patrick Cavanagh.
\newblock The cognitive impenetrability of cognition.
\newblock \emph{Behavioral and Brain Sciences}, 22\penalty0 (3):\penalty0 370--371, 1999.

\bibitem[Chen et~al.(2017)Chen, Zhou, Sun, and Bovik]{chen2017visual}
Jianyu Chen, Jun Zhou, Jun Sun, and Alan~Conrad Bovik.
\newblock Visual discomfort prediction on stereoscopic 3d images without explicit disparities.
\newblock \emph{Signal Processing: Image Communication}, 51:\penalty0 50--60, 2017.

\bibitem[Chen et~al.(2024)Chen, Wan, Matsuda, Ninan, Chapiro, and Sun]{chen2024pea}
Kenneth Chen, Thomas Wan, Nathan Matsuda, Ajit Ninan, Alexandre Chapiro, and Qi Sun.
\newblock Pea-pods: Perceptual evaluation of algorithms for power optimization in xr displays.
\newblock \emph{ACM Transactions on Graphics (TOG)}, 43\penalty0 (4):\penalty0 1--17, 2024.

\bibitem[Chen et~al.(2019)Chen, Jin, Goodall, Yu, and Bovik]{chen2019study}
Meixu Chen, Yize Jin, Todd Goodall, Xiangxu Yu, and Alan~Conrad Bovik.
\newblock Study of 3d virtual reality picture quality.
\newblock \emph{IEEE Journal of Selected Topics in Signal Processing}, 14\penalty0 (1):\penalty0 89--102, 2019.

\bibitem[Chen et~al.(2013{\natexlab{a}})Chen, Cormack, and Bovik]{chen2013no}
Ming-Jun Chen, Lawrence~K Cormack, and Alan~C Bovik.
\newblock No-reference quality assessment of natural stereopairs.
\newblock \emph{IEEE Transactions on Image Processing}, 22\penalty0 (9):\penalty0 3379--3391, 2013{\natexlab{a}}.

\bibitem[Chen et~al.(2013{\natexlab{b}})Chen, Su, Kwon, Cormack, and Bovik]{chen2013full}
Ming-Jun Chen, Che-Chun Su, Do-Kyoung Kwon, Lawrence~K Cormack, and Alan~C Bovik.
\newblock Full-reference quality assessment of stereopairs accounting for rivalry.
\newblock \emph{Signal Processing: Image Communication}, 28\penalty0 (9):\penalty0 1143--1155, 2013{\natexlab{b}}.

\bibitem[Dai et~al.(2024)Dai, Tan, Xu, Futschik, Du, Fanello, Qi, and Zhang]{dai2024svg}
Peng Dai, Feitong Tan, Qiangeng Xu, David Futschik, Ruofei Du, Sean Fanello, Xiaojuan Qi, and Yinda Zhang.
\newblock Svg: 3d stereoscopic video generation via denoising frame matrix.
\newblock \emph{arXiv preprint arXiv:2407.00367}, 2024.

\bibitem[Depthify.ai(2024)]{depthifyDepthifyai}
Depthify.ai.
\newblock Depthify.ai | convert 2d videos to 3d spatial videos.
\newblock \url{http://depthify.ai}, 2024.

\bibitem[Ding et~al.(2018)Ding, Deng, Xie, Xu, Zhao, Chen, and Krylov]{ding2018no}
Yong Ding, Ruizhe Deng, Xin Xie, Xiaogang Xu, Yang Zhao, Xiaodong Chen, and Andrey~S Krylov.
\newblock No-reference stereoscopic image quality assessment using convolutional neural network for adaptive feature extraction.
\newblock \emph{IEEE Access}, 6:\penalty0 37595--37603, 2018.

\bibitem[Dodgson(2004)]{dodgson2004variation}
Neil~A Dodgson.
\newblock Variation and extrema of human interpupillary distance.
\newblock In \emph{Stereoscopic displays and virtual reality systems XI}, pages 36--46. SPIE, 2004.

\bibitem[Dubois(2001)]{dubois2001anaglyph}
Eric Dubois.
\newblock A projection method to generate anaglyph stereo images.
\newblock pages 1661 -- 1664 vol.3, 2001.

\bibitem[Duckworth et~al.(2023)Duckworth, Hedman, Reiser, Zhizhin, Thibert, Lučić, Szeliski, and Barron]{duckworth2023smerf}
Daniel Duckworth, Peter Hedman, Christian Reiser, Peter Zhizhin, Jean-François Thibert, Mario Lučić, Richard Szeliski, and Jonathan~T. Barron.
\newblock Smerf: Streamable memory efficient radiance fields for real-time large-scene exploration, 2023.

\bibitem[Fan et~al.(2024)Fan, Cong, Wen, Wang, Zhang, Ding, Xu, Ivanovic, Pavone, Pavlakos, et~al.]{fan2024instantsplat}
Zhiwen Fan, Wenyan Cong, Kairun Wen, Kevin Wang, Jian Zhang, Xinghao Ding, Danfei Xu, Boris Ivanovic, Marco Pavone, Georgios Pavlakos, et~al.
\newblock Instantsplat: Unbounded sparse-view pose-free gaussian splatting in 40 seconds.
\newblock \emph{arXiv preprint arXiv:2403.20309}, 2024.

\bibitem[Fang et~al.(2019)Fang, Yan, Liu, and Wang]{fang2019stereoscopic}
Yuming Fang, Jiebin Yan, Xuelin Liu, and Jiheng Wang.
\newblock Stereoscopic image quality assessment by deep convolutional neural network.
\newblock \emph{Journal of Visual Communication and Image Representation}, 58:\penalty0 400--406, 2019.

\bibitem[Fu et~al.(2023)Fu, Tamir, Sundaram, Chai, Zhang, Dekel, and Isola]{fu2023dreamsim}
Stephanie Fu, Netanel Tamir, Shobhita Sundaram, Lucy Chai, Richard Zhang, Tali Dekel, and Phillip Isola.
\newblock Dreamsim: Learning new dimensions of human visual similarity using synthetic data.
\newblock \emph{NeurIPS}, 2023.

\bibitem[Guan et~al.(2023)Guan, Penner, Hegland, Letham, and Lanman]{guan2023perceptual}
Phillip Guan, Eric Penner, Joel Hegland, Benjamin Letham, and Douglas Lanman.
\newblock Perceptual requirements for world-locked rendering in ar and vr.
\newblock In \emph{SIGGRAPH Asia 2023 Conference Papers}, pages 1--10, 2023.

\bibitem[Hedman et~al.(2018)Hedman, Philip, Price, Frahm, Drettakis, and Brostow]{DeepBlending2018}
Peter Hedman, Julien Philip, True Price, Jan-Michael Frahm, George Drettakis, and Gabriel Brostow.
\newblock Deep blending for free-viewpoint image-based rendering.
\newblock 37\penalty0 (6):\penalty0 257:1--257:15, 2018.

\bibitem[Hu et~al.(2022)Hu, Shen, Wallis, Allen-Zhu, Li, Wang, Wang, and Chen]{hu2021lora}
Edward~J Hu, Yelong Shen, Phillip Wallis, Zeyuan Allen-Zhu, Yuanzhi Li, Shean Wang, Lu Wang, and Weizhu Chen.
\newblock Lora: Low-rank adaptation of large language models.
\newblock \emph{ICLR}, 2022.

\bibitem[Hua et~al.(2020)Hua, Kohli, Uplavikar, Ravi, Gunaseelan, Orozco, and Li]{hua2020holopix50k}
Yiwen Hua, Puneet Kohli, Pritish Uplavikar, Anand Ravi, Saravana Gunaseelan, Jason Orozco, and Edward Li.
\newblock Holopix50k: A large-scale in-the-wild stereo image dataset.
\newblock In \emph{CVPR Workshop on Computer Vision for Augmented and Virtual Reality, Seattle, WA, 2020.}, 2020.

\bibitem[Ilharco et~al.(2021)Ilharco, Wortsman, Wightman, Gordon, Carlini, Taori, Dave, Shankar, Namkoong, Miller, Hajishirzi, Farhadi, and Schmidt]{ilharco_gabriel_2021_5143773}
Gabriel Ilharco, Mitchell Wortsman, Ross Wightman, Cade Gordon, Nicholas Carlini, Rohan Taori, Achal Dave, Vaishaal Shankar, Hongseok Namkoong, John Miller, Hannaneh Hajishirzi, Ali Farhadi, and Ludwig Schmidt.
\newblock Openclip, 2021.
\newblock If you use this software, please cite it as below.

\bibitem[{Immersity AI}(2024)]{immersity}
{Immersity AI}.
\newblock Immersity ai | convert image and video to 3d.
\newblock \url{https://www.immersity.ai}, 2024.

\bibitem[Insights(2024)]{fortunebusinessinsightsVirtualReality}
Fortune~Business Insights.
\newblock {V}irtual {R}eality [{V}{R}] {M}arket {S}ize, {G}rowth, {S}hare | {R}eport, 2032 --- fortunebusinessinsights.com.
\newblock \url{https://www.fortunebusinessinsights.com/industry-reports/virtual-reality-market-101378}, 2024.
\newblock [Accessed 16-10-2024].

\bibitem[Jain et~al.(2022)Jain, Mildenhall, Barron, Abbeel, and Poole]{jain2022zero}
Ajay Jain, Ben Mildenhall, Jonathan~T Barron, Pieter Abbeel, and Ben Poole.
\newblock Zero-shot text-guided object generation with dream fields.
\newblock In \emph{CVPR}, pages 867--876, 2022.

\bibitem[Jia and Zhang(2018)]{jia2018saliency}
Sen Jia and Yang Zhang.
\newblock Saliency-based deep convolutional neural network for no-reference image quality assessment.
\newblock \emph{Multimedia Tools and Applications}, 77:\penalty0 14859--14872, 2018.

\bibitem[Ke et~al.(2024)Ke, Obukhov, Huang, Metzger, Daudt, and Schindler]{ke2023repurposing}
Bingxin Ke, Anton Obukhov, Shengyu Huang, Nando Metzger, Rodrigo~Caye Daudt, and Konrad Schindler.
\newblock Repurposing diffusion-based image generators for monocular depth estimation.
\newblock In \emph{CVPR}, 2024.

\bibitem[Kerbl et~al.(2023)Kerbl, Kopanas, Leimk{\"u}hler, and Drettakis]{kerbl20233d}
Bernhard Kerbl, Georgios Kopanas, Thomas Leimk{\"u}hler, and George Drettakis.
\newblock 3d gaussian splatting for real-time radiance field rendering.
\newblock \emph{ACM Transactions on Graphics}, 42\penalty0 (4):\penalty0 1--14, 2023.

\bibitem[Kim and Sohn(2011)]{kim2011visual}
Donghyun Kim and Kwanghoon Sohn.
\newblock Visual fatigue prediction for stereoscopic image.
\newblock \emph{IEEE transactions on circuits and systems for video technology}, 21\penalty0 (2):\penalty0 231--236, 2011.

\bibitem[Kim et~al.(2019)Kim, Jeong, taek Lim, and Ro]{Kim2019BinocularFN}
Hak~Gu Kim, Hyunwook Jeong, Heoun taek Lim, and Yong~Man Ro.
\newblock Binocular fusion net: Deep learning visual comfort assessment for stereoscopic 3d.
\newblock \emph{IEEE Transactions on Circuits and Systems for Video Technology}, 29:\penalty0 956--967, 2019.

\bibitem[Knapitsch et~al.(2017)Knapitsch, Park, Zhou, and Koltun]{Knapitsch2017}
Arno Knapitsch, Jaesik Park, Qian-Yi Zhou, and Vladlen Koltun.
\newblock Tanks and temples: Benchmarking large-scale scene reconstruction.
\newblock \emph{ACM Transactions on Graphics}, 36\penalty0 (4), 2017.

\bibitem[Kooi and Toet(2004)]{kooi2004visual}
Frank~L Kooi and Alexander Toet.
\newblock Visual comfort of binocular and 3d displays.
\newblock \emph{Displays}, 25\penalty0 (2-3):\penalty0 99--108, 2004.

\bibitem[Krajancich et~al.(2020)Krajancich, Kellnhofer, and Wetzstein]{10.1145/3414685.3417820}
Brooke Krajancich, Petr Kellnhofer, and Gordon Wetzstein.
\newblock Optimizing depth perception in virtual and augmented reality through gaze-contingent stereo rendering.
\newblock \emph{ACM Trans. Graph.}, 39\penalty0 (6), 2020.

\bibitem[Krizhevsky et~al.(2012)Krizhevsky, Sutskever, and Hinton]{krizhevsky2012imagenet}
Alex Krizhevsky, Ilya Sutskever, and Geoffrey~E Hinton.
\newblock Imagenet classification with deep convolutional neural networks.
\newblock \emph{NeurIPS}, 2012.

\bibitem[Lambooij et~al.(2009)Lambooij, IJsselsteijn, Fortuin, Heynderickx, et~al.]{lambooij2009visual}
Marc Lambooij, Wijnand IJsselsteijn, Marten Fortuin, Ingrid Heynderickx, et~al.
\newblock Visual discomfort and visual fatigue of stereoscopic displays: A review.
\newblock \emph{Journal of imaging science and technology}, 53\penalty0 (3):\penalty0 30201--1, 2009.

\bibitem[LeCun et~al.(1998)LeCun, Bottou, Bengio, and Haffner]{lecun1998gradient}
Yann LeCun, L{\'e}on Bottou, Yoshua Bengio, and Patrick Haffner.
\newblock Gradient-based learning applied to document recognition.
\newblock \emph{Proceedings of the IEEE}, 86\penalty0 (11):\penalty0 2278--2324, 1998.

\bibitem[Li et~al.(2019)Li, Yang, Wan, Wang, Gao, Zhang, and Sun]{li2019no}
Yafei Li, Feng Yang, Wenbo Wan, Jun Wang, Min Gao, Jia Zhang, and Jiande Sun.
\newblock No-reference stereoscopic image quality assessment based on visual attention and perception.
\newblock \emph{IEEE Access}, 7:\penalty0 46706--46716, 2019.

\bibitem[Lim et~al.(2018)Lim, Kim, and Ra]{lim2018vr}
Heaun-Taek Lim, Hak~Gu Kim, and Yang~Man Ra.
\newblock Vr iqa net: Deep virtual reality image quality assessment using adversarial learning.
\newblock In \emph{2018 IEEE International Conference on Acoustics, Speech and Signal Processing (ICASSP)}, pages 6737--6741. IEEE, 2018.

\bibitem[Lin et~al.(2023)Lin, Gao, Tang, Takikawa, Zeng, Huang, Kreis, Fidler, Liu, and Lin]{lin2023magic3d}
Chen-Hsuan Lin, Jun Gao, Luming Tang, Towaki Takikawa, Xiaohui Zeng, Xun Huang, Karsten Kreis, Sanja Fidler, Ming-Yu Liu, and Tsung-Yi Lin.
\newblock Magic3d: High-resolution text-to-3d content creation.
\newblock In \emph{CVPR}, 2023.

\bibitem[Liu et~al.(2019)Liu, Liu, and Shen]{liu2019Learning}
Tsung-Jung Liu, Kuan-Hsien Liu, and Kuan-Hung Shen.
\newblock Learning based no-reference metric for assessing quality of experience of stereoscopic images.
\newblock \emph{Journal of Visual Communication and Image Representation}, 61:\penalty0 272--283, 2019.

\bibitem[Liu et~al.(2020)Liu, Tang, Zheng, and Lin]{liu2020no}
Yun Liu, Chang Tang, Zhi Zheng, and Liyuan Lin.
\newblock No-reference stereoscopic image quality evaluator with segmented monocular features and perceptual binocular features.
\newblock \emph{Neurocomputing}, 405:\penalty0 126--137, 2020.

\bibitem[Lv et~al.(2024)Lv, Long, Huang, Li, Lv, Ren, and Zheng]{lv2024spatialdreamer}
Zhen Lv, Yangqi Long, Congzhentao Huang, Cao Li, Chengfei Lv, Hao Ren, and Dian Zheng.
\newblock Spatialdreamer: Self-supervised stereo video synthesis from monocular input.
\newblock \emph{IEEE VR}, 2024.

\bibitem[Matsuda et~al.(2022)Matsuda, Chapiro, Zhao, Smith, Bachy, and Lanman]{matsuda2022realistic}
Nathan Matsuda, Alex Chapiro, Yang Zhao, Clinton Smith, Romain Bachy, and Douglas Lanman.
\newblock Realistic luminance in vr.
\newblock In \emph{SIGGRAPH Asia 2022 Conference Papers}, pages 1--8, 2022.

\bibitem[Mehl et~al.(2023)Mehl, Schmalfuss, Jahedi, Nalivayko, and Bruhn]{Mehl2023_Spring}
Lukas Mehl, Jenny Schmalfuss, Azin Jahedi, Yaroslava Nalivayko, and Andr\'es Bruhn.
\newblock Spring: A high-resolution high-detail dataset and benchmark for scene flow, optical flow and stereo.
\newblock In \emph{CVPR}, 2023.

\bibitem[Meng et~al.(2022)Meng, He, Song, Song, Wu, Zhu, and Ermon]{meng2021sdedit}
Chenlin Meng, Yutong He, Yang Song, Jiaming Song, Jiajun Wu, Jun-Yan Zhu, and Stefano Ermon.
\newblock Sdedit: Guided image synthesis and editing with stochastic differential equations.
\newblock \emph{ICLR}, 2022.

\bibitem[Menze and Geiger(2015)]{menze2015object}
Moritz Menze and Andreas Geiger.
\newblock Object scene flow for autonomous vehicles.
\newblock In \emph{CVPR}, 2015.

\bibitem[Mildenhall et~al.(2020)Mildenhall, Srinivasan, Tancik, Barron, Ramamoorthi, and Ng]{mildenhall2020nerf}
Ben Mildenhall, Pratul~P. Srinivasan, Matthew Tancik, Jonathan~T. Barron, Ravi Ramamoorthi, and Ren Ng.
\newblock Nerf: Representing scenes as neural radiance fields for view synthesis.
\newblock In \emph{ECCV}, 2020.

\bibitem[Mittal et~al.(2012)Mittal, Moorthy, and Bovik]{mittal2012no}
Anish Mittal, Anush~Krishna Moorthy, and Alan~Conrad Bovik.
\newblock No-reference image quality assessment in the spatial domain.
\newblock \emph{IEEE Transactions on image processing}, 21\penalty0 (12):\penalty0 4695--4708, 2012.

\bibitem[Moorthy et~al.(2013)Moorthy, Su, Mittal, and Bovik]{moorthy2013subjective}
Anush~Krishna Moorthy, Che-Chun Su, Anish Mittal, and Alan~Conrad Bovik.
\newblock Subjective evaluation of stereoscopic image quality.
\newblock \emph{Signal Processing: Image Communication}, 28\penalty0 (8):\penalty0 870--883, 2013.

\bibitem[NRG(2022)]{NRG2022beyond}
NRG.
\newblock Beyond reality: Is the long-awaited vr revolution finally on the horizon?, 2022.

\bibitem[Oh et~al.(2018)Oh, Ahn, Lee, and Bovik]{oh2018deep}
Heeseok Oh, Sewoong Ahn, Sanghoon Lee, and Alan~Conrad Bovik.
\newblock Deep visual discomfort predictor for stereoscopic 3d images.
\newblock \emph{IEEE Transactions on Image Processing}, 27\penalty0 (11):\penalty0 5420--5432, 2018.

\bibitem[Oquab et~al.(2024)Oquab, Darcet, Moutakanni, Vo, Szafraniec, Khalidov, Fernandez, Haziza, Massa, El-Nouby, et~al.]{oquab2023dinov2}
Maxime Oquab, Timoth{\'e}e Darcet, Th{\'e}o Moutakanni, Huy Vo, Marc Szafraniec, Vasil Khalidov, Pierre Fernandez, Daniel Haziza, Francisco Massa, Alaaeldin El-Nouby, et~al.
\newblock Dinov2: Learning robust visual features without supervision.
\newblock \emph{TMLR}, 2024.

\bibitem[Owl3D(2024)]{owl3dOwl3D}
Owl3D.
\newblock Owl3d | ai-powered 2d to 3d conversion software.
\newblock \url{https://www.owl3d.com}, 2024.

\bibitem[Park et~al.(2014)Park, Oh, Lee, and Bovik]{park20143d}
Jincheol Park, Heeseok Oh, Sanghoon Lee, and Alan~Conrad Bovik.
\newblock 3d visual discomfort predictor: Analysis of disparity and neural activity statistics.
\newblock \emph{IEEE transactions on image processing}, 24\penalty0 (3):\penalty0 1101--1114, 2014.

\bibitem[Poole et~al.(2023)Poole, Jain, Barron, and Mildenhall]{poole2022dreamfusion}
Ben Poole, Ajay Jain, Jonathan~T. Barron, and Ben Mildenhall.
\newblock Dreamfusion: Text-to-3d using 2d diffusion.
\newblock \emph{ICLR}, 2023.

\bibitem[Porac and Coren(1976)]{porac1976dominant}
Clare Porac and Stanley Coren.
\newblock The dominant eye.
\newblock \emph{Psychological bulletin}, 83\penalty0 (5):\penalty0 880, 1976.

\bibitem[Poreddy et~al.(2023)Poreddy, Ganeswaram, Appina, Kokil, and Pachori]{poreddy2023no}
Ajay Kumar~Reddy Poreddy, Raja Bharath~Chandra Ganeswaram, Balasubramanyam Appina, Priyanka Kokil, and Ram~Bilas Pachori.
\newblock No-reference virtual reality image quality evaluator using global and local natural scene statistics.
\newblock \emph{IEEE Transactions on Instrumentation and Measurement}, 2023.

\bibitem[Poulakos et~al.(2015)Poulakos, Monroy, Aydin, Wang, Smolic, and Gross]{poulakos2015computational}
Steven Poulakos, Rafael Monroy, Tunc Aydin, Oliver Wang, Aljoscha Smolic, and Markus Gross.
\newblock A computational model for perception of stereoscopic window violations.
\newblock In \emph{2015 Seventh International Workshop on Quality of Multimedia Experience (QoMEX)}, pages 1--6. IEEE, 2015.

\bibitem[Qi et~al.(2012)Qi, Jiang, Ma, and Zhao]{qi2012Quality}
Feng Qi, Tingting Jiang, Siwei Ma, and Debin Zhao.
\newblock Quality of experience assessment for stereoscopic images.
\newblock In \emph{2012 IEEE International Symposium on Circuits and Systems (ISCAS)}, pages 1712--1715, 2012.

\bibitem[Radford et~al.(2021)Radford, Kim, Hallacy, Ramesh, Goh, Agarwal, Sastry, Askell, Mishkin, Clark, et~al.]{radford2021learning}
Alec Radford, Jong~Wook Kim, Chris Hallacy, Aditya Ramesh, Gabriel Goh, Sandhini Agarwal, Girish Sastry, Amanda Askell, Pamela Mishkin, Jack Clark, et~al.
\newblock Learning transferable visual models from natural language supervision.
\newblock In \emph{ICML}, 2021.

\bibitem[Ranftl et~al.(2022)Ranftl, Lasinger, Hafner, Schindler, and Koltun]{Ranftl2022}
Ren\'{e} Ranftl, Katrin Lasinger, David Hafner, Konrad Schindler, and Vladlen Koltun.
\newblock Towards robust monocular depth estimation: Mixing datasets for zero-shot cross-dataset transfer.
\newblock \emph{IEEE Transactions on Pattern Analysis and Machine Intelligence}, 44\penalty0 (3), 2022.

\bibitem[Rehm(2018)]{dxomarkHYDROGENHolographic}
Lars Rehm.
\newblock {R}{E}{D} {H}{Y}{D}{R}{O}{G}{E}{N} {H}olographic {S}martphone - first look --- dxomark.com.
\newblock \url{https://www.dxomark.com/red-hydrogen-holographic-smartphone-first-look/}, 2018.
\newblock [Accessed 25-11-2024].

\bibitem[Rubini(2023)]{treendlyCapturingFuture}
Mike Rubini.
\newblock {C}apturing the {F}uture: {E}xploring the {E}ver-{I}ncreasing {P}opularity of {D}ual {C}amera {S}martphones --- treendly.com.
\newblock \url{https://treendly.com/blog/capturing-the-future-exploring-the-ever-increasing-popularity-of-dual-camera-smartphones}, 2023.
\newblock [Accessed 25-11-2024].

\bibitem[Scharstein et~al.(2014)Scharstein, Hirschm{\"u}ller, Kitajima, Krathwohl, Ne{\v{s}}i{\'c}, Wang, and Westling]{scharstein2014high}
Daniel Scharstein, Heiko Hirschm{\"u}ller, York Kitajima, Greg Krathwohl, Nera Ne{\v{s}}i{\'c}, Xi Wang, and Porter Westling.
\newblock High-resolution stereo datasets with subpixel-accurate ground truth.
\newblock In \emph{GCPR}, 2014.

\bibitem[Shah(2018)]{counterpointresearchDualCamera}
Neil Shah.
\newblock {D}ual {C}amera {S}martphones {A}re {G}oing {M}ainstream: {E}xplosive {G}rowth in 2018 --- counterpointresearch.com.
\newblock \url{https://www.counterpointresearch.com/insights/dual-camera-smartphones-going-mainstream-explosive-growth-2018/}, 2018.
\newblock [Accessed 16-10-2024].

\bibitem[Shen et~al.(2018)Shen, Fang, Yao, Geng, and Wu]{shen2018no}
Liquan Shen, Ruigang Fang, Yang Yao, Xianqiu Geng, and Dapeng Wu.
\newblock No-reference stereoscopic image quality assessment based on image distortion and stereo perceptual information.
\newblock \emph{IEEE Transactions on Emerging Topics in Computational Intelligence}, 3\penalty0 (1):\penalty0 59--72, 2018.

\bibitem[Shen et~al.(2021)Shen, Chen, Pan, Fan, Li, and Lei]{shen2021no}
Lili Shen, Xiongfei Chen, Zhaoqing Pan, Kefeng Fan, Fei Li, and Jianjun Lei.
\newblock No-reference stereoscopic image quality assessment based on global and local content characteristics.
\newblock \emph{Neurocomputing}, 424:\penalty0 132--142, 2021.

\bibitem[Shi et~al.(2024)Shi, Li, and Wonka]{shi2024immersepro}
Jian Shi, Zhenyu Li, and Peter Wonka.
\newblock Immersepro: End-to-end stereo video synthesis via implicit disparity learning.
\newblock \emph{arXiv preprint arXiv:2410.00262}, 2024.

\bibitem[Shibata et~al.(2011)Shibata, Kim, Hoffman, and Banks]{shibata2011zone}
Takashi Shibata, Joohwan Kim, David~M Hoffman, and Martin~S Banks.
\newblock The zone of comfort: Predicting visual discomfort with stereo displays.
\newblock \emph{Journal of vision}, 11\penalty0 (8):\penalty0 11--11, 2011.

\bibitem[Si et~al.(2022)Si, Huang, Yang, Lin, and Pan]{si2022no}
Jianwei Si, Baoxiang Huang, Huan Yang, Weisi Lin, and Zhenkuan Pan.
\newblock A no-reference stereoscopic image quality assessment network based on binocular interaction and fusion mechanisms.
\newblock \emph{IEEE Transactions on Image Processing}, 31:\penalty0 3066--3080, 2022.

\bibitem[Stokes(2013)]{stokes2013cognitive}
Dustin Stokes.
\newblock Cognitive penetrability of perception.
\newblock \emph{Philosophy Compass}, 8\penalty0 (7):\penalty0 646--663, 2013.

\bibitem[Sui et~al.(2021)Sui, Ma, Yao, and Fang]{sui2021perceptual}
Xiangjie Sui, Kede Ma, Yiru Yao, and Yuming Fang.
\newblock Perceptual quality assessment of omnidirectional images as moving camera videos.
\newblock \emph{IEEE Transactions on Visualization and Computer Graphics}, 28\penalty0 (8):\penalty0 3022--3034, 2021.

\bibitem[Suvorov et~al.(2022)Suvorov, Logacheva, Mashikhin, Remizova, Ashukha, Silvestrov, Kong, Goka, Park, and Lempitsky]{suvorov2022resolution}
Roman Suvorov, Elizaveta Logacheva, Anton Mashikhin, Anastasia Remizova, Arsenii Ashukha, Aleksei Silvestrov, Naejin Kong, Harshith Goka, Kiwoong Park, and Victor Lempitsky.
\newblock Resolution-robust large mask inpainting with fourier convolutions.
\newblock In \emph{WACV}, pages 2149--2159, 2022.

\bibitem[Thomas(2020)]{thomas2020examining}
Bruce~H Thomas.
\newblock Examining user perception of the size of multiple objects in virtual reality.
\newblock \emph{Applied Sciences}, 10\penalty0 (11):\penalty0 4049, 2020.

\bibitem[Ukai and Howarth(2008)]{ukai2008visual}
Kazuhiko Ukai and Peter~A Howarth.
\newblock Visual fatigue caused by viewing stereoscopic motion images: Background, theories, and observations.
\newblock \emph{Displays}, 29\penalty0 (2):\penalty0 106--116, 2008.

\bibitem[Wang et~al.(2014)Wang, Zeng, and Wang]{wang2014quality}
Jiheng Wang, Kai Zeng, and Zhou Wang.
\newblock Quality prediction of asymmetrically distorted stereoscopic images from single views.
\newblock In \emph{2014 IEEE International Conference on Multimedia and Expo (ICME)}, pages 1--6. IEEE, 2014.

\bibitem[Wang et~al.(2015)Wang, Rehman, Zeng, Wang, and Wang]{wang2015quality}
Jiheng Wang, Abdul Rehman, Kai Zeng, Shiqi Wang, and Zhou Wang.
\newblock Quality prediction of asymmetrically distorted stereoscopic 3d images.
\newblock \emph{IEEE Transactions on Image Processing}, 24\penalty0 (11):\penalty0 3400--3414, 2015.

\bibitem[Wang et~al.(2023{\natexlab{a}})Wang, Chan, and Loy]{wang2022exploring}
Jianyi Wang, Kelvin~CK Chan, and Chen~Change Loy.
\newblock Exploring clip for assessing the look and feel of images.
\newblock In \emph{AAAI}, 2023{\natexlab{a}}.

\bibitem[Wang et~al.(2023{\natexlab{b}})Wang, Yuan, Chen, Zhang, Wang, and Zhang]{wang2023modelscope}
Jiuniu Wang, Hangjie Yuan, Dayou Chen, Yingya Zhang, Xiang Wang, and Shiwei Zhang.
\newblock Modelscope text-to-video technical report.
\newblock \emph{arXiv preprint arXiv:2308.06571}, 2023{\natexlab{b}}.

\bibitem[Wang et~al.(2019)Wang, Wang, Yang, An, and Guo]{Flickr1024}
Yingqian Wang, Longguang Wang, Jungang Yang, Wei An, and Yulan Guo.
\newblock Flickr1024: A large-scale dataset for stereo image super-resolution.
\newblock In \emph{International Conference on Computer Vision Workshops}, pages 3852--3857, 2019.

\bibitem[Wang et~al.(2024{\natexlab{a}})Wang, Lu, Wang, Bao, Li, Su, and Zhu]{wang2024prolificdreamer}
Zhengyi Wang, Cheng Lu, Yikai Wang, Fan Bao, Chongxuan Li, Hang Su, and Jun Zhu.
\newblock Prolificdreamer: High-fidelity and diverse text-to-3d generation with variational score distillation.
\newblock \emph{NeurIPS}, 36, 2024{\natexlab{a}}.

\bibitem[Wang et~al.(2024{\natexlab{b}})Wang, Yuan, Wang, Chen, Xia, Luo, and Shan]{wang2023motionctrl}
Zhouxia Wang, Ziyang Yuan, Xintao Wang, Tianshui Chen, Menghan Xia, Ping Luo, and Yin Shan.
\newblock Motionctrl: A unified and flexible motion controller for video generation.
\newblock 2024{\natexlab{b}}.

\bibitem[Weinzaepfel et~al.(2022)Weinzaepfel, Leroy, Lucas, Brégier, Cabon, Arora, Antsfeld, Chidlovskii, Csurka, and Revaud]{weinzaepfel2023crocoselfsupervisedpretraining3d}
Philippe Weinzaepfel, Vincent Leroy, Thomas Lucas, Romain Brégier, Yohann Cabon, Vaibhav Arora, Leonid Antsfeld, Boris Chidlovskii, Gabriela Csurka, and Jérôme Revaud.
\newblock Croco: Self-supervised pre-training for 3d vision tasks by cross-view completion, 2022.

\bibitem[Weinzaepfel et~al.(2023)Weinzaepfel, Lucas, Leroy, Cabon, Arora, Brégier, Csurka, Antsfeld, Chidlovskii, and Revaud]{weinzaepfel2023crocov2improvedcrossview}
Philippe Weinzaepfel, Thomas Lucas, Vincent Leroy, Yohann Cabon, Vaibhav Arora, Romain Brégier, Gabriela Csurka, Leonid Antsfeld, Boris Chidlovskii, and Jérôme Revaud.
\newblock Croco v2: Improved cross-view completion pre-training for stereo matching and optical flow, 2023.

\bibitem[Wu et~al.(2023)Wu, Zhang, Zhang, Chen, Liao, Li, Gao, Wang, Zhang, Sun, et~al.]{wu2023q}
Haoning Wu, Zicheng Zhang, Weixia Zhang, Chaofeng Chen, Liang Liao, Chunyi Li, Yixuan Gao, Annan Wang, Erli Zhang, Wenxiu Sun, et~al.
\newblock Q-align: Teaching lmms for visual scoring via discrete text-defined levels.
\newblock \emph{arXiv preprint arXiv:2312.17090}, 2023.

\bibitem[Yang et~al.(2024{\natexlab{a}})Yang, Kang, Huang, Xu, Feng, and Zhao]{yang2024depth}
Lihe Yang, Bingyi Kang, Zilong Huang, Xiaogang Xu, Jiashi Feng, and Hengshuang Zhao.
\newblock Depth anything: Unleashing the power of large-scale unlabeled data.
\newblock In \emph{CVPR}, 2024{\natexlab{a}}.

\bibitem[Yang et~al.(2022)Yang, Wu, Shi, Lao, Gong, Cao, Wang, and Yang]{yang2022maniqa}
Sidi Yang, Tianhe Wu, Shuwei Shi, Shanshan Lao, Yuan Gong, Mingdeng Cao, Jiahao Wang, and Yujiu Yang.
\newblock Maniqa: Multi-dimension attention network for no-reference image quality assessment.
\newblock In \emph{CVPR}, 2022.

\bibitem[Yang et~al.(2024{\natexlab{b}})Yang, Teng, Zheng, Ding, Huang, Xu, Yang, Hong, Zhang, Feng, et~al.]{yang2024cogvideox}
Zhuoyi Yang, Jiayan Teng, Wendi Zheng, Ming Ding, Shiyu Huang, Jiazheng Xu, Yuanming Yang, Wenyi Hong, Xiaohan Zhang, Guanyu Feng, et~al.
\newblock Cogvideox: Text-to-video diffusion models with an expert transformer.
\newblock \emph{arXiv preprint arXiv:2408.06072}, 2024{\natexlab{b}}.

\bibitem[Yano et~al.(2004)Yano, Emoto, and Mitsuhashi]{yano2004two}
Sumio Yano, Masaki Emoto, and Tetsuo Mitsuhashi.
\newblock Two factors in visual fatigue caused by stereoscopic hdtv images.
\newblock \emph{Displays}, 25\penalty0 (4):\penalty0 141--150, 2004.

\bibitem[Yi et~al.(2024)Yi, Fang, Wang, Wu, Xie, Zhang, Liu, Tian, and Wang]{yi2024gaussiandreamer}
Taoran Yi, Jiemin Fang, Junjie Wang, Guanjun Wu, Lingxi Xie, Xiaopeng Zhang, Wenyu Liu, Qi Tian, and Xinggang Wang.
\newblock Gaussiandreamer: Fast generation from text to 3d gaussians by bridging 2d and 3d diffusion models.
\newblock In \emph{CVPR}, pages 6796--6807, 2024.

\bibitem[Yu et~al.(2024)Yu, Chen, Huang, Sattler, and Geiger]{yu2024mip}
Zehao Yu, Anpei Chen, Binbin Huang, Torsten Sattler, and Andreas Geiger.
\newblock Mip-splatting: Alias-free 3d gaussian splatting.
\newblock In \emph{CVPR}, 2024.

\bibitem[Zhang et~al.(2023)Zhang, Li, and Chang]{zhang2023towards}
Huilin Zhang, Sumei Li, and Yongli Chang.
\newblock Towards top-down stereoscopic image quality assessment via stereo attention.
\newblock \emph{arXiv preprint arXiv:2308.04156}, 2023.

\bibitem[Zhang et~al.(2018)Zhang, Isola, Efros, Shechtman, and Wang]{zhang2018unreasonable}
Richard Zhang, Phillip Isola, Alexei~A Efros, Eli Shechtman, and Oliver Wang.
\newblock The unreasonable effectiveness of deep features as a perceptual metric.
\newblock In \emph{CVPR}, 2018.

\bibitem[Zhang et~al.(2016)Zhang, Qu, Ma, Guan, and Huang]{zhang2016learning}
Wei Zhang, Chenfei Qu, Lin Ma, Jingwei Guan, and Rui Huang.
\newblock Learning structure of stereoscopic image for no-reference quality assessment with convolutional neural network.
\newblock \emph{Pattern Recognition}, 59:\penalty0 176--187, 2016.

\bibitem[Zhao et~al.(2024)Zhao, Hu, Cun, Zhang, Li, Kong, Gao, Niu, and Shan]{zhao2024stereocrafter}
Sijie Zhao, Wenbo Hu, Xiaodong Cun, Yong Zhang, Xiaoyu Li, Zhe Kong, Xiangjun Gao, Muyao Niu, and Ying Shan.
\newblock Stereocrafter: Diffusion-based generation of long and high-fidelity stereoscopic 3d from monocular videos.
\newblock \emph{arXiv preprint arXiv:2409.07447}, 2024.

\bibitem[Zhou et~al.(2019)Zhou, Chen, and Li]{zhou2019dual}
Wei Zhou, Zhibo Chen, and Weiping Li.
\newblock Dual-stream interactive networks for no-reference stereoscopic image quality assessment.
\newblock \emph{IEEE Transactions on Image Processing}, 28\penalty0 (8):\penalty0 3946--3958, 2019.

\bibitem[Zhou et~al.(2023)Zhou, Chen, Yin, Huang, and Li]{zhou2023stereoscopic}
Yang Zhou, Pingan Chen, Haibing Yin, Xiaofeng Huang, and Zhu Li.
\newblock Stereoscopic image discomfort prediction using dual-stream multi-level interactive network.
\newblock \emph{Displays}, 78:\penalty0 102444, 2023.

\bibitem[Zhu et~al.(2024)Zhu, Yang, Duan, Min, Zhai, and Callet]{zhu2024esiqa}
Xilei Zhu, Liu Yang, Huiyu Duan, Xiongkuo Min, Guangtao Zhai, and Patrick~Le Callet.
\newblock Esiqa: Perceptual quality assessment of vision-pro-based egocentric spatial images.
\newblock \emph{arXiv preprint arXiv:2407.21363}, 2024.

\end{thebibliography}
}

\clearpage
\setcounter{page}{1}
\maketitlesupplementary

\section{Dataset Details}
\label{sec:dataset_breakdown}
\begin{figure*}[t]
    \centering
    \includegraphics[width=0.9\linewidth]{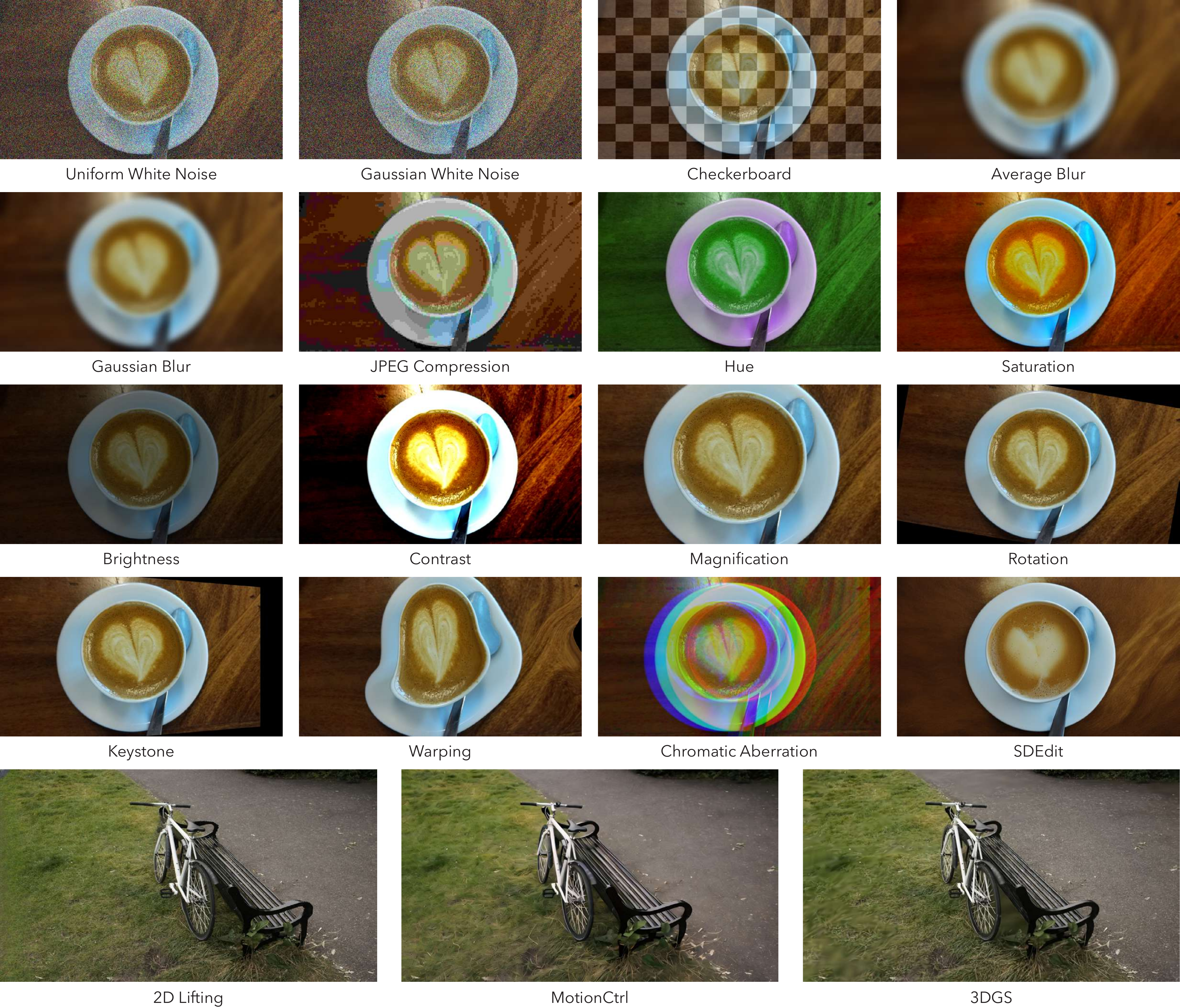}
    \caption{\small{\bf Distortion examples.} We show examples of image distortions in our dataset, exaggerated for illustration purposes.}
    \label{fig:all_augmentations}
    \vspace{-2mm}
\end{figure*}

Our dataset is comprised of $2400$ examples, each containing a pair of stereo images, resulting in a total of $4800$ stereo images, all of which have undergone some form of distortion. \Cref{tab:augmentation_numbers} presents the total amount in which each distortion appears in the dataset. \Cref{fig:all_augmentations} shows visual examples for several distortions, exaggerated for illustration purposes.

\begin{table}[ht]
    \begin{center}
    \begin{tabular}{ r l }
    \toprule
    \textbf{Distortion type} & \textbf{Occurrence} \\ 
    \midrule
    2D lifting & $1364$ \\
    \midrule
    MotionCtrl & $1156$ \\
    \midrule
    3D Gaussian splatting & $306$ \\
    \midrule
    SDEdit & $241$ \\
    \midrule
    Uniform White Noise & $152$ \\
    \midrule
    Chromatic Aberration & $144$ \\
    \midrule
    Rotation & $141$ \\
    \midrule
    Keystone & $138$ \\
    \midrule
    Average Blur & $127$ \\
    \midrule
    Gaussian Blur & $125$ \\
    \midrule
    JPEG Compression & $124$ \\
    \midrule
    Gaussian White Noise & $123$ \\
    \midrule
    Checkerboard & $117$ \\
    \midrule
    Warping & $115$ \\
    \midrule
    Brightness & $97$ \\
    \midrule
    Saturation & $95$ \\
    \midrule
    Contrast & $80$ \\
    \midrule
    Hue & $79$ \\
    \midrule
    Magnification & $76$ \\
    \bottomrule
    \end{tabular}
    \caption{\small \textbf{Frequency of applied distortions in our proposed dataset.}}
    \label{tab:augmentation_numbers}   
    \end{center}
\end{table}

\section{Training Details}
\label{sec:sm_training_details}
Our model is trained on a single NVIDIA A100 using an Adam optimizer with a learning rate of $3e-5$ and batch size of $16$. We maintain the original $1280 \times 720$ resolution, only applying a center crop to $1274 \times 714$ for compatibility with DINOv2-S's patch size of $14$. During training, we finetune the DINOv2 backbone using LoRA, with a rank of $8$, alpha of $32$, and dropout of $0.1$. We use a with a margin of $0.05$ for the hinge loss. Additionally, the training data is weighted based on annotator consensus levels, with each epoch taking approximately $12$ minutes to train and $1$ minute to validate. 

\section{Detailed Performance on the SCOPE dataset}
\label{sec:sm_performance}

\begin{figure*}[ht]
    \centering
    \includegraphics[width=0.95\linewidth]{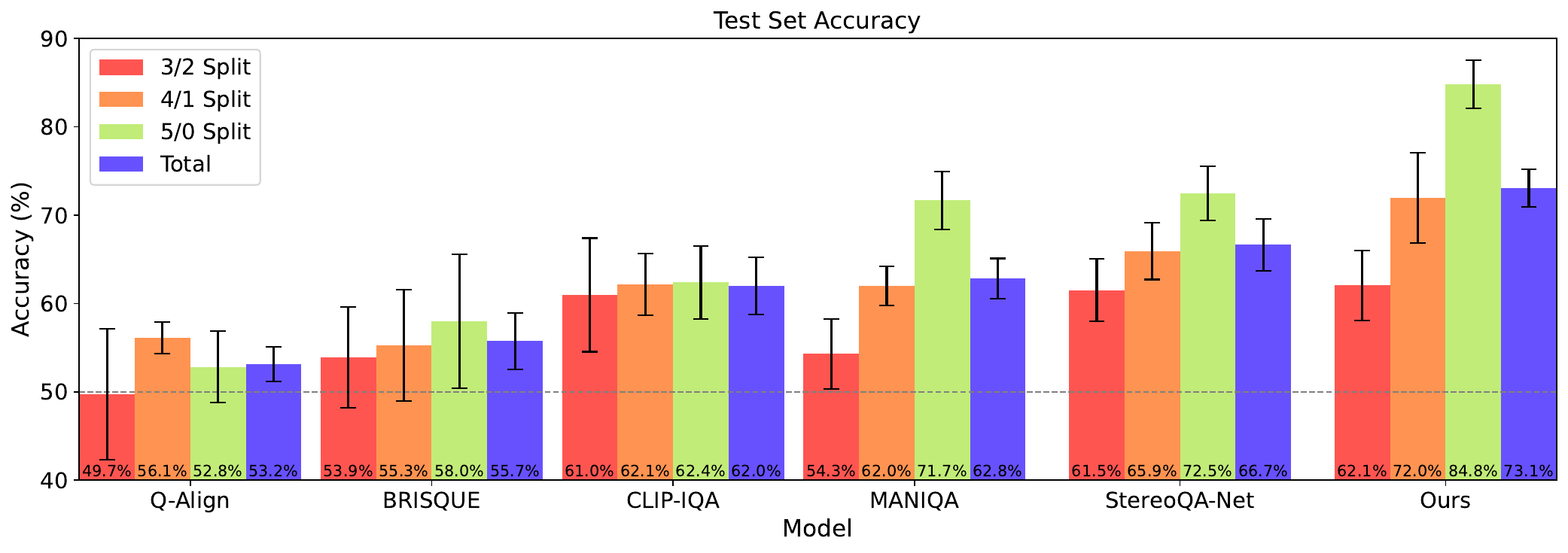}
    \caption{\small{\bf Test set performance.} We test the performance of existing IQA and NR-SIQA models as well as our proposed model on a held out test sets, and show the results on different splits.}
    \label{fig:performance}
\end{figure*}

In \Cref{fig:performance} we report test set accuracy across several different train, validation and test partitions, categorized by annotation consensus: unanimous ($5-0$ split), majority ($4-1$ split), and divided ($3-2$ split), with the latter being the noisiest and most cognitively penetrable. The sizes of these splits are similar, comprising $32.9\%$, $34.1\%$, and $32.9\%$ of the data respectively, confirming that our dataset contains a learnable signal. We train and evaluate our model on five $80\%-10\%-10\%$ dataset splits, using five different seeds for each split, and report the mean and standard deviation in \Cref{fig:performance}.

\begin{figure*}[ht]
    \centering
    \includegraphics[width=1\linewidth]{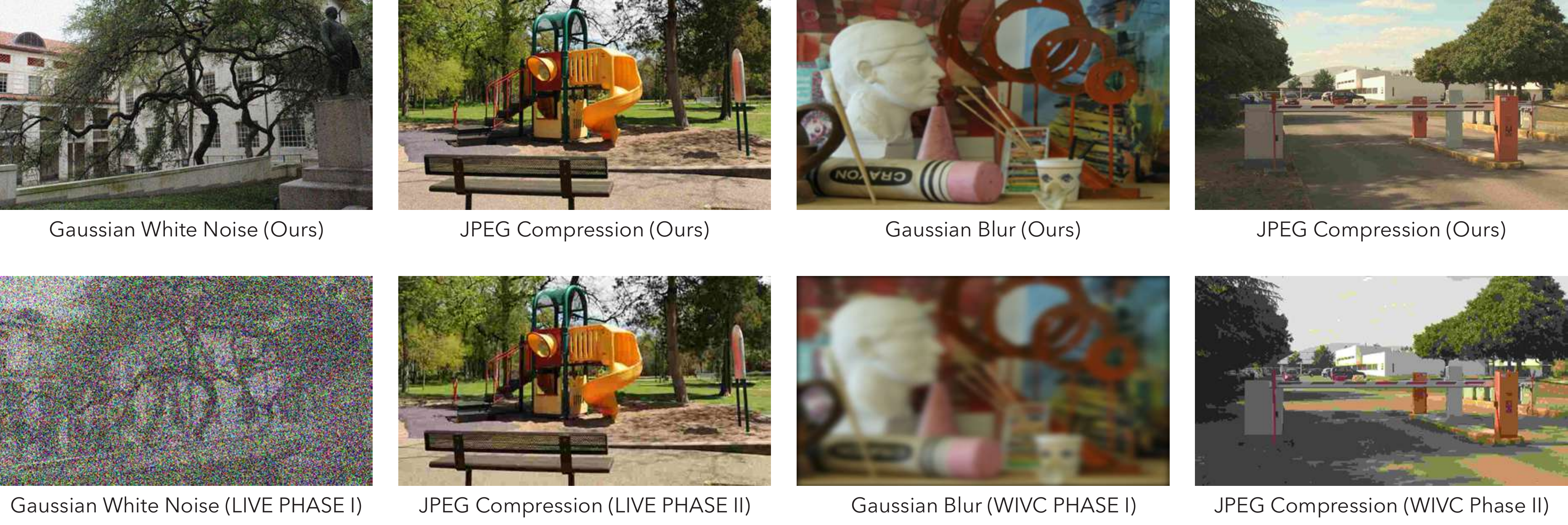}
    \caption{\small{\bf Distortion strength comparison.} Comparing maximum distortion strengths across existing datasets and our proposed dataset, demonstrating that the distortions applied in our dataset exhibit significantly lower intensity compared to existing SIQA datasets.}
    \label{fig:other_datasets}
\end{figure*}

\begin{table*}[ht]
\begin{center}
\resizebox{1\textwidth}{!}{
\begin{tabular}{c c cc cc cc cc}
 \toprule
 \multicolumn{1}{c}{} & \multicolumn{1}{c}{} & \multicolumn{2}{c}{\bf LIVE Phase I} & \multicolumn{2}{c}{\bf LIVE Phase II} & \multicolumn{2}{c}{\bf WIVC Phase I} & \multicolumn{2}{c}{\bf WIVC Phase II} \\
 \cmidrule(lr){3-4}\cmidrule(lr){5-6}\cmidrule(lr){7-8}\cmidrule(lr){9-10}
 {\bf} & {\bf Method} & {\bf SROCC $\uparrow$} & {\bf PLCC $\uparrow$} & {\bf SROCC $\uparrow$} & {\bf PLCC $\uparrow$}  & {\bf SROCC $\uparrow$} & {\bf PLCC $\uparrow$} & {\bf SROCC $\uparrow$} & {\bf PLCC $\uparrow$} \\ 
 \midrule
 \multirow{4}{*}{\shortstack[c]{\bf Manual \\\textbf{Feature} \\\textbf{Based}}} 
 & Chen \etal \cite{chen2013no} & $0.891$ & $0.895$ & $0.880$ & $0.880$ & -- & -- & -- & --\\
 & Shen \etal \cite{shen2018no} & $0.932$ & $0.936$ & $0.927$ & $0.932$ & -- & -- & -- & --\\
  & Li \etal \cite{li2019no} & $0.953$ & $0.965$ & $0.946$ & $0.955$ & $0.937$ & $0.949$ & $0.952$ & $0.960$\\
   & Liu \etal \cite{liu2020no} & $0.949$ & $0.958$ & $0.933$ & $0.935$ & $0.928$ & $0.945$ & $0.901$ & $0.913$\\
 \midrule
 \multirow{7}{*}{\shortstack[c]{\bf Deep \\\textbf{Learning} \\\textbf{Based}}} 
 & Zhang \etal \cite{zhang2016learning} & $0.943$ & $0.947$ & $0.915$ & $0.912$ & -- & -- & -- & --\\
 & Ding \etal \cite{ding2018no} & $0.942$ & $0.940$ & $0.924$ & $0.930$ & -- & -- & -- & --\\
  & Fang \etal \cite{fang2019stereoscopic} & $0.946$ & $0.957$ & $0.934$ & $0.946$ & -- & -- & -- & --\\
 & Zhou \etal \cite{zhou2019dual} & $0.965$ & $0.973$ & $0.947$ & $0.957$ & -- & --  & -- & --\\
 & Shen \etal \cite{shen2021no} & $0.962$ & $0.972$ & $0.951$ & $0.953$ & -- & -- & -- & --\\
 & Si \etal \cite{si2022no} & $0.966$ & $0.978$ & $0.953$ & $0.972$ & $0.960$ & $0.969$ & $0.950$ & $0.958$ \\ 
 & Zhang \etal \cite{zhang2023towards} & $0.972$ & $0.977$ & $0.962$ & $0.964$ & $0.972$ & $0.973$ & $0.972$ & $0.973$ \\
 \midrule
& iSQoE (Ours) & 0.774 & 0.758 & 0.763 & 0.767 & 0.627 & 0.687 & 0.542 & 0.536 \\
 \bottomrule
\end{tabular}
}
\caption{\small {\bf Evaluation on existing datasets for stereoscopic image quality assessment.}}\label{tab:siqa_results_full}
\end{center}
\end{table*}

\newpage
\section{Performance on Existing SIQA Datasets}
\label{sec:siqa_evaluation}

\Cref{tab:stereoscopic_datasets} provides a comparison between our dataset and existing stereo quality assessment datasets: LIVE 3D Phases I and II~\cite{moorthy2013subjective,chen2013full}, Waterloo IVC (WIVC) 3D Phases I and II~\cite{wang2014quality,wang2015quality} and IEEE-SA~\cite{liu2019Learning}. SCOPE differs from them in several aspects:

\begin{enumerate}[leftmargin=15pt]
    \item \textbf{Image Quantity:} SCOPE is the largest of the datasets, with more than twice the amount of samples than IEEE-SA - the second largest dataset.
    \item \textbf{Annotation medium:} The annotations in all these datasets were collected using passive stereoscopic displays or active shutter glasses, while ours were collected on a Vision Pro headset. In \Cref{sec:cross_device} and \Cref{fig:kappa} we demonstrate low correlation between preferences on VR devices and other stereo viewing methods.
    \item \textbf{Annotation Protocol:} The other datasets collected Mean Opinion Score annotations, an absolute single-image protocol, while SCOPE collected 2AFC which are relative annotations.
    \item \textbf{Distortion Strengths:} The other datasets applied significantly stronger distortions than SCOPE, see \Cref{fig:other_datasets}.
\end{enumerate}

We evaluate our model on LIVE 3D Phase I and II~\cite{moorthy2013subjective,chen2013no,chen2013full} and Waterloo IVC (WIVC) 3D Phase I and II~\cite{wang2014quality,wang2015quality}. For these evaluations, we use standard performance metrics: Spearman rank order correlation coefficient (SROCC) and Pearson linear correlation coefficient (PLCC).

\Cref{tab:siqa_results_full} shows there is a significant performance gap between our models and the state-of-the-art models reporting performance on these datasets. We attribute this to the difference in annotation mediums between these datasets and SCOPE. Our model is trained and fitted to grade quality as it is perceived on a VR device, rather than on passive stereoscopic displays.

\begin{figure}[ht]
    \centering
    \includegraphics[width=1.0\linewidth]{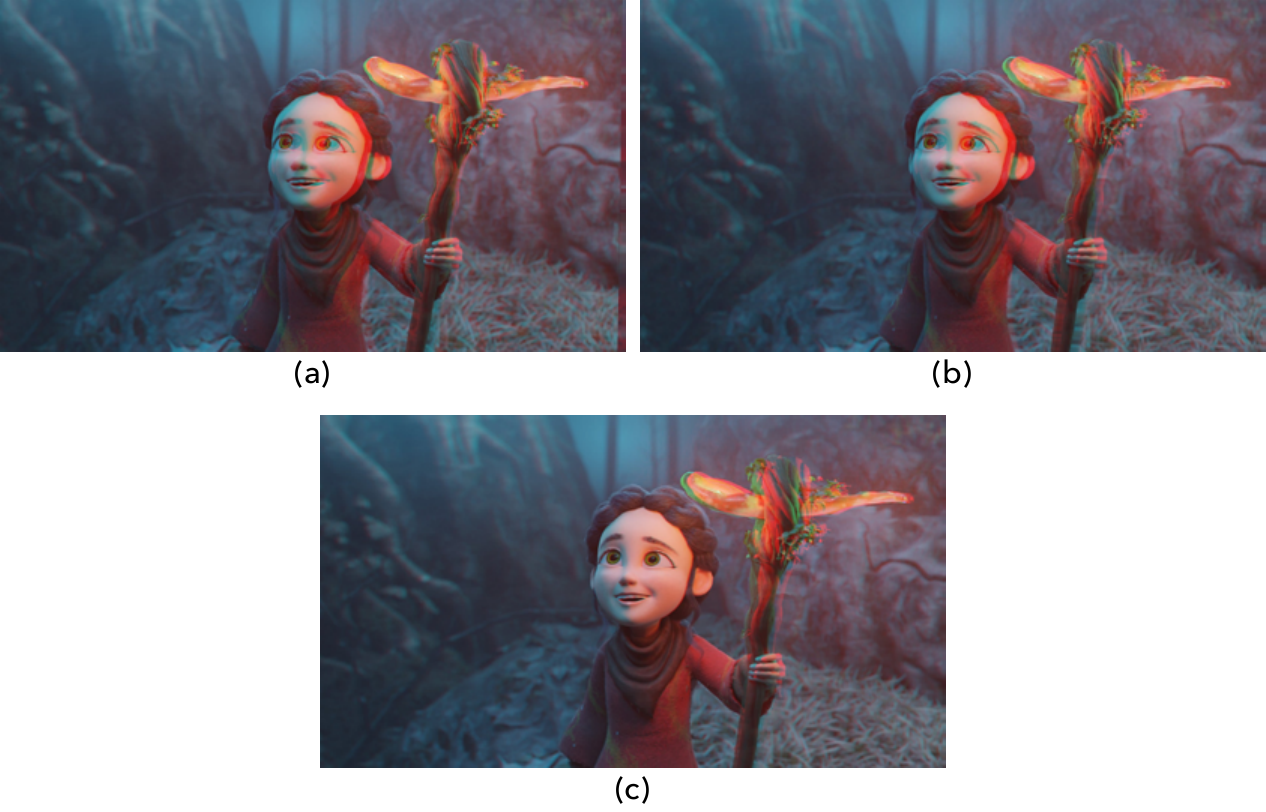}
      \caption{\small{\bf Example from our off-the-shelf, mono-to-stereo experiment.} Three versions of the same stereo image are generated using different off-the-shelf, mono-to-stereo conversion methods. (a) Depthify.ai (b) Immersity AI (c) Owl3D. The stereo images are presented as anaglyph images for viewing purposes. We recommend viewing the images on a screen and zooming in to better observe the differences.}
    \label{fig:spring_examples}
\end{figure}

\begin{table}[t]
    \begin{center}
    \resizebox{1.0\linewidth}{!}{
    \begin{tabular}{lccccc}
    \toprule
    \multirow{2}{*}{\textbf{Dataset}} & \multirow{2}{*}{\textbf{Samples}} & \textbf{Stereo} & \textbf{Clean} & \textbf{Annotation} & \multirow{2}{*}{\textbf{Distortions}}
    \\
    & & \textbf{Images} & \textbf{Images} & \textbf{Type} &
    \\
    \midrule
    LIVE & \multirow{3}{*}{$365$} & \multirow{3}{*}{$365$} & \multirow{3}{*}{$20$} & \multirow{3}{*}{DMOS} & Noise, Blur, 
    \\
    Phase I & & & & & Compression,
    \\
    \cite{moorthy2013subjective} & & & & & Fast-fading
    \\
    \midrule
    LIVE & \multirow{3}{*}{$360$} & \multirow{3}{*}{$360$} & \multirow{3}{*}{$8$} & \multirow{3}{*}{DMOS} & Noise, Blur,
    \\
    Phase II & & & & & Compression,
    \\
    \cite{chen2013full} & & & & & Fast-fading
    \\
    \midrule
    WIVC & \multirow{3}{*}{$330$} & \multirow{3}{*}{$330$} & \multirow{3}{*}{$6$} & \multirow{3}{*}{MOS} & \multirow{3}{*}{Noise, Blur}
    \\
    Phase I & & & & &
    \\
    \cite{wang2014quality} & & & & &
    \\
    \midrule
    WIVC & \multirow{3}{*}{$460$} & \multirow{3}{*}{$460$} & \multirow{3}{*}{$10$} & \multirow{3}{*}{MOS} & Noise, Blur,
    \\
    Phase II & & & & & Compression,
    \\
    \cite{wang2015quality} & & & & &
    \\
    \midrule
    \multirow{2}{*}{IEEE-SA~\cite{liu2019Learning}} & \multirow{2}{*}{$800$} & \multirow{2}{*}{$800$} & \multirow{2}{*}{$160$} & \multirow{2}{*}{MOS} & Horizontal
    \\
    & & & & & disparity
    \\
    \midrule
    SCOPE & \multirow{2}{*}{$2400$} & \multirow{2}{*}{$4800$} & \multirow{2}{*}{$2400$} & \multirow{2}{*}{2AFC} & $19$ types, 
    \\
    (Ours) & & & & & see \Cref{tab:augmentations}
    \\
    \bottomrule
    \end{tabular}
    }
    \caption{\small \textbf{Stereoscopic Preference Datasets.} Prior datasets for stereo image evaluation vary in terms of size, the psychophysical experiment in which the annotations were collected, and the distortions they encompass.}
    \label{tab:stereoscopic_datasets}
    \end{center}
    \vspace{-7mm}
\end{table}

\begin{figure*}[ht]
    \centering
    \includegraphics[width=0.95\linewidth]{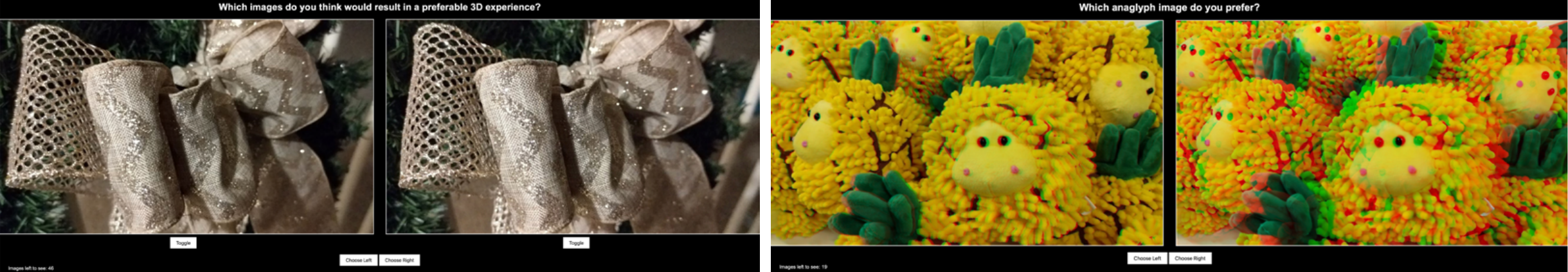}
    \caption{\small{\bf User study setups: left/right image toggling (top) and anaglyph stereo (bottom)}}    
    \label{fig:html_pages}
\end{figure*}

\newpage
\section{Cross-Medium User Study}
\label{sec:cross_device_sm}
Expanding on the user study outlined in Section 4.3, we detail the specific viewing setups for each device. Viewing stereoscopic images with the Apple Vision Pro was done through the native photos app in immersive mode. For the Meta Quest Pro, we employed a third-party application (Pegasus VR media player) due to the absence of a suitable first-party viewer. Both the toggling and anaglyph setups were presented via HTML pages, shown in \Cref{fig:html_pages}. We opted for full-color anaglyph images, as this convention provided the best stereoscopic 3D experience with our monitor and glasses combination, among all conventions tested.

\begin{figure*}[ht]
    \centering
    \includegraphics[width=0.8\linewidth]{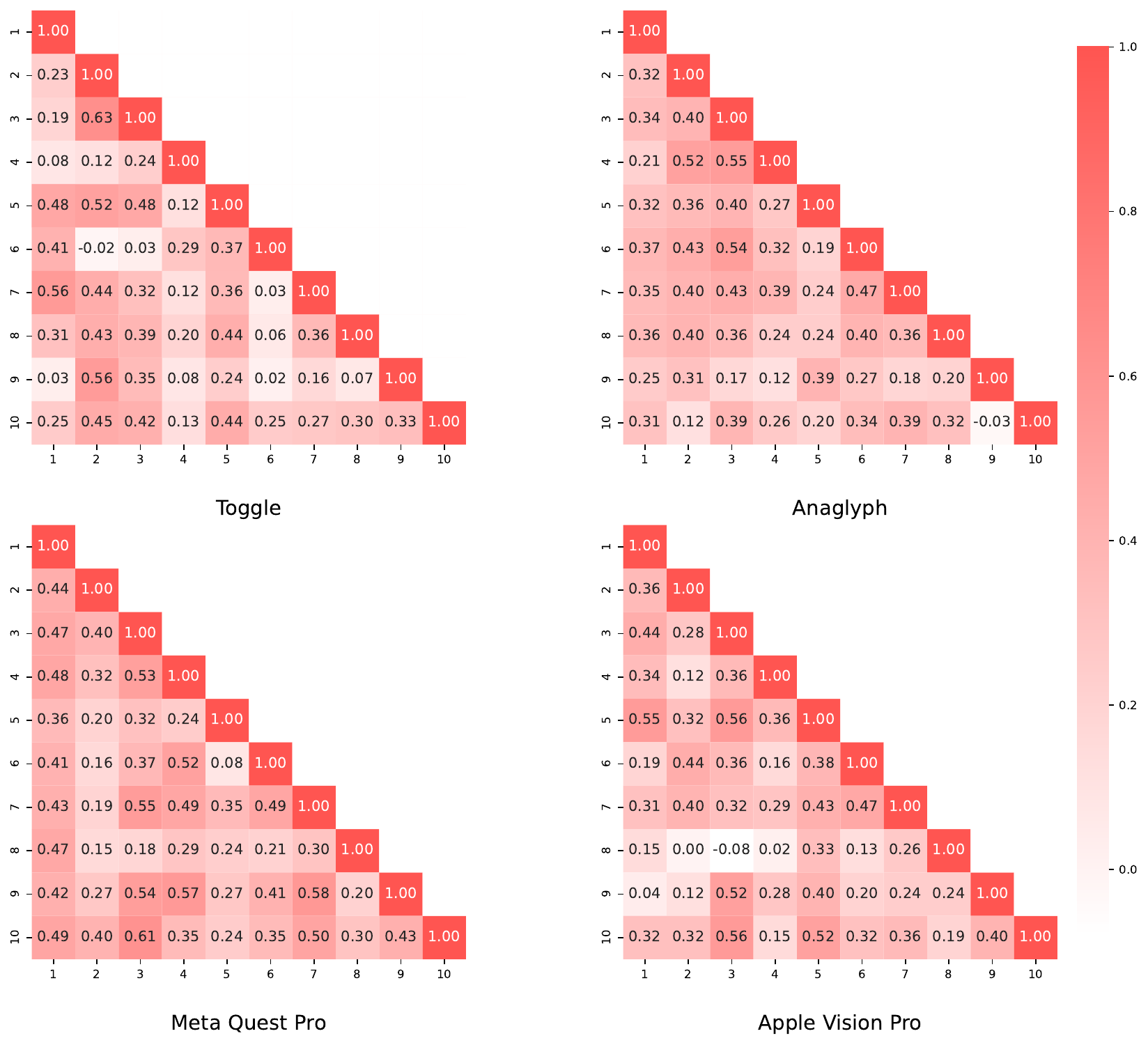}
    \caption{\small{\bf Inter-rater agreement for each viewing medium, measured using Cohen’s kappa coefficient.} The heatmap displays the agreement scores between all pairs of the 10 participants, highlighting the correlation in subjective evaluations across different viewing conditions.}
    \label{fig:agreement}
\end{figure*}

In addition to \Cref{fig:kappa} that shows the mean correlation, \Cref{fig:agreement} we show the Cohen's kappa coefficient between each of the 10 participants for each viewing device. 

\section{Off-the-Shelf Mono-to-Stereo Evaluation}
\label{sec:mono-to-stereo-evaluation}

We evaluated alignment of human opinion with the different SQoE candidates on the Spring~\cite{Mehl2023_Spring} dataset. \Cref{fig:spring_examples} shows an example from the user study.

\section{Licenses}
The models and datasets we use are provided under the licenses in \Cref{tab:licenses}.

\begin{table}[ht]
\resizebox{0.45\textwidth}{!}{
\begin{tabular}{l l | l l}
\toprule
\multicolumn{1}{l}{\textbf{Dataset}} & \multicolumn{1}{l|}{\textbf{License}} & \multicolumn{1}{l}{\textbf{Model}} & \multicolumn{1}{l}{\textbf{License}} \\ \midrule
 Tanks and Temples & CC BY 4.0 & MotionCtrl & Apache 2.0 \\
 Deep Blending & Apache 2.0 & MiDaS & MIT \\
 Mip-NeRF 360 & Apache 2.0 & Marigold & Apache 2.0 \\
 Holopix50k & NC & Depth Anything & Apache 2.0 \\
 SPRING & CC BY 4.0 & LaMa & Apache 2.0 \\
 LIVE 3D Phase I & Custom Academic & 3DGS & NC \\
 LIVE 3D Phase II & Custom Academic & DINOv2 & Apache 2.0 \\
 WIVC 3D Phase I & Custom Academic & Q-Align & S-Lab 1.0 \\
 WIVC 3D Phase II & Custom Academic & BRISQUE & Apache 2.0 \\
 & & CLIP-IQA & S-Lab 1.0\\
 & & MANIQA & Apache 2.0 \\
 & & StereoQA-Net & Custom Academic \\
 & & CLIP & MIT \\
 & & OpenCLIP & MIT \\
 & & Croco & CC BY-NC-SA 4.0 \\
 & & Depthify.ai & Custom \\
 & & Immersity AI & Custom \\
 & & Owl3D & Custom \\ \bottomrule
\end{tabular}}
\caption{\small{\bf Dataset and model licenses.}}
\label{tab:licenses}
\end{table}

\end{document}